\documentclass{ieeeaccess}
\usepackage{etex}
\usepackage{cite}
\usepackage{amsmath,amssymb,amsfonts}
\usepackage{graphicx,color}
\usepackage{textcomp}
\usepackage{xcolor}
\usepackage{hyperref}
\hypersetup{hidelinks}
\usepackage[compatibility=false]{caption}
\usepackage{subcaption}
\usepackage{placeins}
\usepackage{booktabs}
\usepackage{mathptmx}
\usepackage{etoolbox}
\newcommand{\boldstep}[1]{\smallskip\noindent\textbf{#1:}}

\newenvironment{myitemize}
  {\begin{list}{$\bullet$}{%
      \setlength{\leftmargin}{1em}%
      \setlength{\labelwidth}{1.5em}
      \setlength{\labelsep}{0.5em}%
      \setlength{\itemindent}{0pt}%
      \setlength{\listparindent}{0pt}%
      \setlength{\topsep}{0.5em}%
      \setlength{\parsep}{0pt}%
      \setlength{\itemsep}{0.2em}%
    }}
  {\end{list}}

\begin{document}
\history{Date of publication xxxx 00, 0000, date of current version xxxx 00, 0000.}
\doi{10.1109/ACCESS.2017.DOI}

\title{Lossy Event Compression: From Event Stream Distortion to Task Performance}

\author{\uppercase{Zahra Rezaee}\authorrefmark{1},
\uppercase{Catarina Brites}\authorrefmark{2}, \IEEEmembership{Member, IEEE},
\uppercase{and Jo\~{a}o Ascenso}\authorrefmark{1}, \IEEEmembership{Senior Member, IEEE}}

\address[1]{Instituto Superior T\'{e}cnico, University of Lisbon, Instituto de Telecomunica\c{c}\~{o}es, Lisbon, Portugal}
\address[2]{Instituto Universit\'{a}rio de Lisboa (ISCTE-IUL), Instituto de Telecomunica\c{c}\~{o}es, Lisbon, Portugal}

\tfootnote{This work is funded by FCT/MECI through national funds and when applicable co-funded EU funds under UID/50008: Instituto de Telecomunicações and the project NeuroVision, https://doi.org/10.54499/2024.17187.PEX. This work has been submitted to the IEEE for possible publication. Copyright may be transferred without notice, after which this version may no longer be accessible.}

\markboth
{Rezaee \headeretal: Lossy Event Compression: From Event Stream Distortion to Task Performance}
{Rezaee \headeretal: Lossy Event Compression: From Event Stream Distortion to Task Performance}

\corresp{Corresponding author: Jo\~{a}o Ascenso (e-mail: joao.ascenso@lx.it.pt).}
\begin{abstract}
Event cameras generate asynchronous, sparse data streams with microsecond temporal resolution, but in moderate-to-high motion scenes they can produce as many as hundreds of millions of events per second, creating significant bandwidth and storage challenges. Lossy compression is therefore essential for practical deployment, yet existing event stream distortion metrics fail to reliably predict compression-induced degradation at the task level, forcing codec optimization to rely on expensive task-specific evaluations. To address this gap, this paper introduces two fundamentally different event compression pipelines: i) an aggregation-based pipeline that converts the event stream into polarity-based histogram frames for compression with the conventional image codec JPEG 2000, and ii) a frame-free point cloud-based pipeline that codes events natively as 3D points using the octree-based codec G-PCC. Both pipelines are then assessed within a unified task-driven evaluation framework that relates event stream distortion to downstream application performance across four representative tasks: i) video reconstruction, ii) object detection,  iii) optical flow estimation, and a delay-sensitive task iv) asynchronous feature tracking under a reference-relative protocol. Building on this framework, five classification-based distortion metrics are applied to event compression for the first time, to the best of the authors' knowledge, and benchmarked against existing event stream metrics. Experimental results demonstrate that the proposed metrics reliably predict compression-induced task degradation across different coding frameworks. This demonstrates that event stream distortion assessment can be an efficient alternative to repeated task-specific evaluation, providing direct guidance for the development and optimization of future event data coding solutions.
\end{abstract}

\titlepgskip=-15pt
\maketitle
\section{Introduction}
\label{sec:introduction}

\PARstart{C}{onventional} frame-based cameras sample the light intensity at all sensor pixel locations simultaneously at fixed time intervals, producing a sequence of dense (2D) spatial frames regardless of scene activity. This synchronous sampling leads to redundant data in static/low-motion environments, motion blur during fast movement, and limited dynamic range in high-contrast lighting conditions. Event-based sensors, also known as neuromorphic vision sensors, address these limitations by mimicking the operation of biological retinas. Rather than acquiring entire frames at fixed time intervals, each pixel independently monitors brightness changes and asynchronously generates a so-called \textit{event} only when the relative brightness change exceeds a threshold. An event is represented as a tuple $(x, y, t, p)$, where $(x, y)$ denotes the pixel coordinates, $t$ is the timestamp with microsecond precision and $p$ indicates the polarity, being positive for brightness increase and negative for decrease. This bio-inspired design of event cameras produces sparse data streams with very high temporal resolution while providing low latency and wide dynamic range, making them well suited for applications such as robotics, autonomous driving and augmented reality~\cite{Gallego:2022}. Importantly, features such as microsecond latency, a dynamic range far exceeding that of conventional sensors and an output whose bandwidth scales with scene activity rather than a fixed frame rate, are intrinsic to this new event sensing paradigm and, therefore, cannot be achieved by conventional frame-based sensors regardless of their frame rate. This is one of the key factors driving the growing interest in event cameras.

\begin{figure}[!htb]
\centering
\begin{minipage}[c]{0.32\linewidth}
\centering
\includegraphics[width=\textwidth, height=3.0cm, keepaspectratio=false]{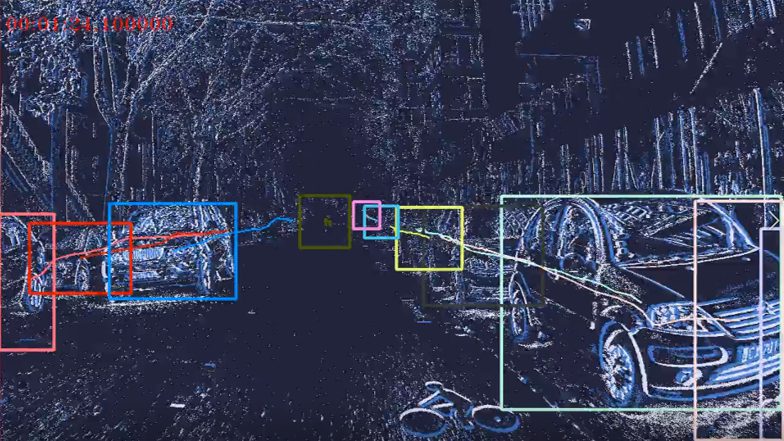}
\end{minipage}
\hfill
\begin{minipage}[c]{0.32\linewidth}
\centering
\includegraphics[width=\textwidth, height=3.0cm, keepaspectratio=false]{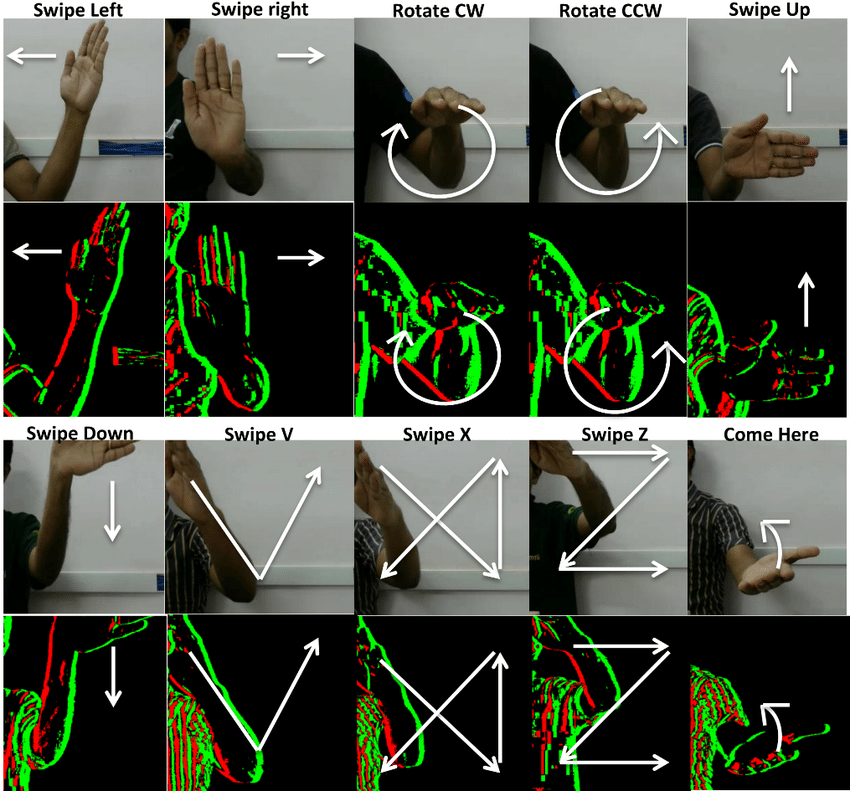}
\end{minipage}
\hfill
\begin{minipage}[c]{0.32\linewidth}
\centering
\includegraphics[width=\textwidth, height=3.0cm, keepaspectratio=false]{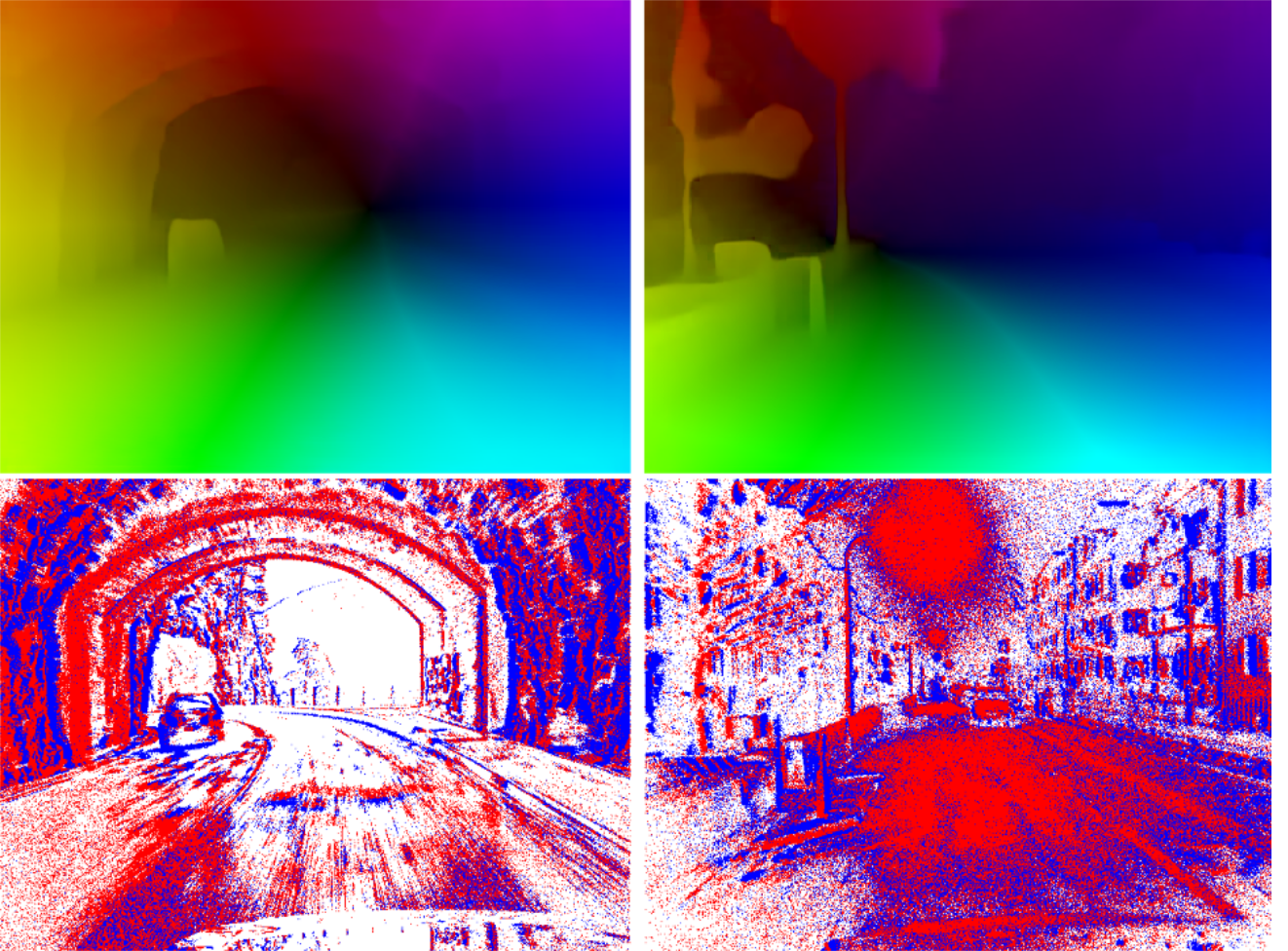}
\end{minipage}
\caption{Some applications of event data: object detection for autonomous driving \cite{GehrigM:2023} (left), gesture recognition for human-computer interaction \cite{Baby:2018} (middle), and optical flow estimation for motion analysis \cite{GehrigM:2021} (right).}
\label{fig:applications}
\end{figure}

However, these cameras can generate tens or even hundreds of millions of events per second, especially for visual scenes with moderate to high motion, which leads to very high bandwidth and storage demands. The compression of event data, also referred to in the literature as neuromorphic vision data coding (NVDC), is currently an active research topic and the recent JPEG exploration activity on event-based vision, known as JPEG XE~\cite{JPEGXE:2023usecases, JPEGXE:2024cfp}, highlights its practical relevance. The scope of JPEG XE is the \textit{creation and development of a standard to represent events in an efficient way allowing interoperability between sensing, storage, and processing, targeting machine vision and other relevant applications}. JPEG XE is expected to become the first international standard for coding of events and, while lossless coding (phases 1--2) has been the main target so far, lossy coding (phase 3) is planned to be addressed soon. However, it has already been recognized by the JPEG group that it is difficult to assess the impact of coding loss and that new evaluation metrics may need to be designed. Over the past years, several NVDC methods have been proposed for both lossless and lossy scenarios, employing diverse coding strategies and evaluation methodologies~\cite{Brites:2025}. According to ~\cite{Brites:2025}, it is essential not only to develop more efficient compression solutions but also to define meaningful performance evaluation methodologies, capable of reflecting the main event data application requirements.

Compressing event streams poses several challenges~\cite{Brites:2025}, particularly when the compressed data must support both downstream machine vision processing and human visualization. Fig.~\ref{fig:applications} illustrates representative applications of event data. In machine vision, event streams are used to extract task-relevant information for applications such as tracking, localization, mapping, and industrial inspection. In human visualization, event data enables improving visual quality in tasks such as image and video deblurring, high dynamic range imaging, slow-motion reconstruction, and augmented reality. However, for lossy event compression, reliable and widely accepted metrics for measuring the impact of coding distortions on application-level performance are still lacking. This challenge  is further intensified by the absence of comprehensive evaluation studies, standard benchmark datasets, and established best practices~\cite{Brites:2025}. In this context, the main objective of this paper is to determine whether intrinsic event stream distortion measures can reliably characterize the impact of lossy compression on downstream task performance. Accordingly, this paper advances the state-of-the-art in event data compression and its systematic evaluation through the following key contributions:
\begin{myitemize}
    \item \textbf{Modular aggregation-based coding framework leveraging image codecs}: Event streams are temporally aggregated into polarity-based 2D histograms (count frames), enabling efficient compression with classical image codecs, followed by reconstruction back into the event domain. The key novelty lies in the proposed inverse aggregation mechanism, which recovers the event stream in a form compatible with any downstream application. While the proposed framework is modular and can accommodate different conventional image codecs, this work adopts the wavelet-based JPEG~2000 codec. This choice is motivated by a comparative evaluation of representative image codecs relying on different transform coding approaches on the video reconstruction task.
    \item \textbf{Frame-free point cloud (PC) based coding framework}: A second, fundamentally different framework encodes the event stream natively as a 3D point cloud in $(x,y,t)$ space using the MPEG geometry-based point cloud codec (G-PCC), preserving a finer-grained temporal representation than the millisecond-order aggregation windows of the first framework, without ever forming an intermediate frame representation. This framework represents events fundamentally differently from the first, using 3D point sets instead of histogram frames, thereby allowing the proposed metrics to be evaluated for sensitivity to the underlying representation rather than to the distortion itself.
    \item \textbf{Task-driven evaluation framework for event data compression}: Compression performance is assessed on representative downstream tasks, including video reconstruction, object detection, and optical flow estimation, covering both human visualization and machine vision use cases. Unlike prior work, the evaluation is conducted across multiple tasks with rather different objectives and characteristics.
    \item \textbf{Classification-based event stream distortion metrics}: A set of classification-based distortion metrics is applied to event data compression for the first time, to the best of the authors' knowledge, and systematically evaluated against well-known benchmarks by analyzing their correlation with task-specific utility measures across multiple compression settings and downstream tasks. This comprehensive analysis identifies the metrics that most reliably predict downstream task degradation. The proposed metrics are after also validated on delay-sensitive task, asynchronous feature tracking where there are much stricter temporal requirements.
\end{myitemize}

Overall, these contributions provide a comprehensive framework for lossy event data compression. By explicitly relating compression-induced distortion to downstream application performance, this work paves the way for application-aware design and optimization of future event compression solutions.

\section{Related Work}\label{sec:background}
Lossy event compression methods reduce data rates by discarding information considered less critical for downstream applications. Due to its practical importance, several lossy compression strategies have been explored in the literature~\cite{Brites:2025}. For example, temporal aggregation-based strategies accumulate events, over fixed time intervals, into a frame-based representation, converting the asynchronous event stream into structured 2D frames that can be processed by standard image or video codecs. The frame-based representation is adopted, for instance, by the TALVEN solution~\cite{Khan:2021}, where events are first aggregated at each pixel location according to their polarity, generating two polarity-based event frames, and HEVC encoding is then applied over event frames. The work by Schiopu and Bilcu~\cite{Schiopu:2022} also considers the compression of (groups of) event frames but adopts a context-based image compression solution. It is worth noting that, although the effective encoding stage may be lossless in both approaches, temporal aggregation inherently introduces timestamp quantization, making the overall compression pipeline lossy. Alternative lossy compression strategies include event sampling methods that selectively retain events based on priority maps~\cite{Banerjee:2021} and predictive coding approaches that transmit optical flow to enable event prediction at the receiver~\cite{stumpp2024flow}.

Point cloud-based strategies, in contrast, represent the event stream directly as a three-dimensional point set, exploiting the natural mapping of the $(x, y, t)$ event coordinates onto a point cloud geometry so that standard point cloud codecs can be applied. This representation is adopted, for instance, by Martini \textit{et al.}~\cite{martini2022lossless}, where the positive and negative events are arranged in 3D space as two separate point clouds, one per polarity, that are then losslessly compressed. The work by Huang and Ebrahimi~\cite{huang2023eventpcc} also addresses lossless compression by encoding the event stream as a point cloud with the MPEG geometry-based codec G-PCC~\cite{gpcc:tmc13} and comparing different strategies for creating the point cloud, including treating the polarity as a point attribute or splitting the events into separate point clouds according to polarity. However, point cloud coding can also operate in a lossy regime by coarsely quantizing the spatial and temporal coordinates. For example, Huang \textit{et al.}~\cite{huang2023lossyeval} apply G-PCC in lossy mode at increasing quantization levels and study how the resulting distortion (e.g. event loss) degrades the accuracy of several computer vision tasks. Adhuran \textit{et al.}~\cite{adhuran2025lossy} also employ lossy G-PCC, but first aggregate the events into two point clouds, one for each polarity, with coordinates representing the pixel position and aggregation interval index, while the corresponding event counts are stored as point attributes; they further evaluate the spatial and temporal errors introduced by the lossy coding using two dedicated fidelity metrics proposed by the authors. 

The works discussed above illustrate the diversity of compression strategies proposed for event data; the reader may refer to~\cite{Brites:2025} for a comprehensive overview of (lossy and lossless) NVDC solutions in the literature. Despite this diversity, existing works mainly evaluate performance in terms of compression ratio and bitrate, without systematically assessing the compression impact on downstream vision tasks. This limitation is particularly relevant given event data is exploited for both machine vision applications and human visual interpretation, each requiring evaluation methods that accurately reflect the compression impact on their specific objectives. In machine vision applications, event streams may be processed by tasks such as object detection, tracking, SLAM and optical flow estimation, where compression errors can directly degrade the task performance by removing informative events or introducing spurious ones. In human-oriented applications, including video reconstruction, deblurring and high dynamic range imaging, the same compression errors can reduce the visual fidelity and interpretability of the reconstructed content. This highlights the need for evaluation methods that are able to explicitly and meaningfully relate event-level compression distortions to their impact on application-level performance, in both machine vision and human visualization scenarios.

While several task-independent event stream distortion metrics have been proposed in the literature, they do not explicitly account for the impact of compression-induced distortions on downstream task performance. In total, six metrics were identified from the literature and are considered here. The Asynchronous Spatiotemporal Spike Metric (ASTSM) originally proposed by Li \textit{et al.}~\cite{Li:2023} and later employed by Stumpp \textit{et al.}~\cite{stumpp2024flow} for event compression evaluation, evaluates the similarity between event streams by measuring their distance in a reproducing Kernel Hilbert space (RKHS) \cite{stumpp2024flow}. To assess geometric accuracy, PSNR Event-to-Event (PSNR E2E) calculates the symmetric Mean Squared Error (MSE) of Euclidean distances in $(x, y, t)$ space, considering both geometry and polarity \cite{seleem2026deep}. For time-aggregated data, Spatial Fidelity ($VQ_s$) utilizes a 2D PSNR metric derived from Euclidean distances found via Approximate Nearest Neighbor (ANN) search, while Temporal Fidelity ($VQ_t$) computes the Root Mean Square (RMS) error of timestamps between original events and their nearest neighbors on the decoded stream \cite{adhuran2025lossy}. Frame PSNR evaluates spatial quality by aggregating events into polarity-based histograms and computing peak signal-to-noise ratio between original frame (obtained from input event stream aggregation) and reconstructed frame (obtained from reconstructed event stream aggregation), while frame MSE reports the corresponding mean squared error~\cite{banerjee2021lossy}.

As none of the aforementioned objective metrics has been evaluated in terms of their ability to predict downstream task performance, it remains an open question which metrics reliably capture functionally relevant distortions in event data. This paper addresses this gap by systematically evaluating the impact of compression across multiple downstream tasks for several existing state-of-the-art metrics and proposing, for the first time, the use of the more efficient classification-based metrics for event stream compression.

\section{Aggregation-Based Event Stream Compression Pipeline}\label{sec:codec}

\begin{figure*}[t]\centering
\includegraphics[width=0.85\textwidth]{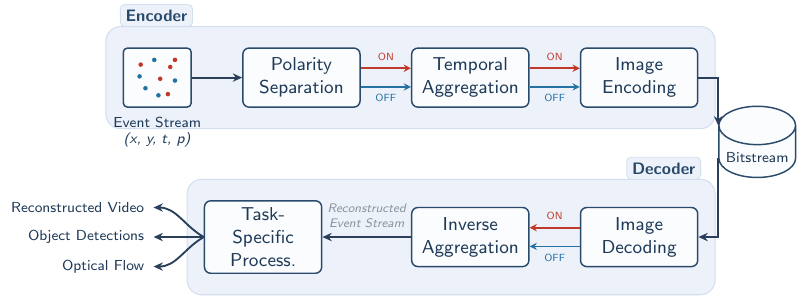}
\caption{Aggregation-based event stream compression pipeline.}\label{fig:agg-codec-pipeline}
\end{figure*}

Figure~\ref{fig:agg-codec-pipeline} illustrates the first of the two event-based compression pipelines considered in this work, which builds on prior temporal aggregation-based approaches~\cite{Khan:2021}~\cite{Schiopu:2022}. This pipeline employs a temporal aggregation-based approach to convert asynchronous event streams into a frame-based representation suitable for standard 2D image compression. The aggregation-based event compression pipeline consists of several key steps. First, polarity-based temporal aggregation converts raw events into histogram (count) frames, which are then lossless or lossy compressed by an image encoder. At the decoder/receiver side, the histogram (count) frames are first reconstructed by an image decoder, followed by inverse aggregation, to reconstruct the event stream from the decoded (histogram) frames. Finally, the reconstructed event stream is fed to the task-specific processing module for downstream task evaluation.
\subsection{Codec Architecture and Walkthrough}\label{ssec:codec_arch}

In the proposed aggregation-based event compression solution, the following operations are performed:

\boldstep{Polarity Separation and Temporal Aggregation} The temporal aggregation stage converts the asynchronous event stream into a structured frame-based representation suitable for image compression. Raw events, represented as 4-tuples $(x, y, t, p)$ containing spatial coordinates, timestamp, and polarity, are first split by polarity into positive (ON) and negative (OFF) event streams and then independently partitioned into fixed-duration time windows. Within each fixed-duration time window, also referred to as the temporal aggregation interval $\Delta t$, events are accumulated into two separate histogram (count) frames, one for each polarity. At every pixel location $(x, y)$, each histogram frame stores the total count of events of the corresponding polarity. The maximum event count per pixel across all temporal aggregation intervals is then determined to select the bit depth needed for the image codec. Given the sparsity of event camera data, a bit depth of 8 bits was sufficient for all datasets and temporal aggregation intervals evaluated in this paper, although a 16-bit representation may be required for scenes with higher event rates or longer temporal aggregation intervals. The temporal aggregation process produces a sequence of grayscale frame pairs that preserve spatial event distributions while aggregating temporal information within each aggregation interval $\Delta t$. 

\boldstep{Image Encoding and Decoding} The two histogram (count) frames obtained in the previous step are independently compressed using JPEG~2000\cite{JPEG2000}, which supports both lossless and lossy compression modes. JPEG 2000 is chosen due to its superior compression efficiency for sparse data, its effective exploitation of spatial redundancy via wavelet decomposition, and its precise rate control through quantization parameters. A more detailed rationale for this choice is provided in Section \ref{ssec:codec_impact}. For lossless compression, the JPEG 2000 codec operates in reversible mode using the 5/3 reversible wavelet transform, ensuring bit-exact reconstruction of histogram (count) frames. For lossy compression, the 9/7 irreversible wavelet transform is employed with variable quantization step sizes to achieve target bitrates. In the decoding stage the histogram frames are reconstructed from the compressed bitstream.

\boldstep{Inverse Aggregation} The inverse aggregation stage converts the decoded histogram (count) frames back into an event stream representation suitable for downstream processing. For each pixel $(x, y)$ with event count $N$ in the time window $[t_{\text{start}}, t_{\text{start}} + \Delta t)$, $N$ events are generated by assigning uniformly spaced timestamps over the time window. This deterministic timestamp assignment strategy eliminates variability in inter-event intervals, resulting in event reconstruction with constant temporal spacing of $\Delta t / N$. Specifically, for $N > 0$, the $N$ reconstructed event timestamps at pixel $(x,y)$ are defined as $t_i = t_{\text{start}} + i \cdot \Delta t / N$ for $i = 0, 1, \ldots, N-1$; pixels with $N = 0$ generate no reconstructed events. Since the original intra-window temporal information is discarded during aggregation, the reassignment of timestamps introduces a temporal quantization error bounded by the aggregation interval $\Delta t$; the reconstructed timestamps retain the sensor's microsecond temporal resolution, but their exact positions within each interval are approximated by the uniform temporal grid. Each reconstructed event is assigned the polarity of its corresponding decoded histogram frame, either positive or negative. After processing all pixels in both decoded polarity frames, the reconstructed events are merged and sorted by timestamp to produce the output (reconstructed) event stream in the usual $(x, y, t, p)$ format. Note that, even with perfect reconstruction of the event counts, the original temporal ordering of events within each aggregation interval cannot be guaranteed to be preserved, since the aggregation process discards the individual event timestamps and retains only their counts per pixel and polarity.
\subsection{Codec Operating Points}\label{ssec:codec_points}
The rate-distortion operating points of the proposed aggregation-based event codec are determined by two configuration parameters:
\begin{myitemize}
\item \textbf{Temporal aggregation interval:} Defines the duration $\Delta t$ of each aggregation window and, consequently, the number of histogram frames generated per second, thereby directly influencing both the bitrate and the temporal fidelity of the reconstructed event stream. Larger temporal aggregation intervals generally result in greater temporal distortion, as more events are accumulated within each histogram frame while their original timestamps are discarded.
\item \textbf{JPEG 2000 target bitrate:} Defines the target bit budget used by post-compression rate–distortion optimization to select the coding pass truncation points that minimize distortion while satisfying the bitrate constraint. This parameter controls the compression level applied to the histogram frames and, consequently, the accuracy of the reconstructed event counts and the spatial fidelity of the reconstructed event stream. Lower target bitrates generally result in larger reconstruction errors, leading to changes in both the total number of reconstructed events and the spatial distribution of event counts. 
\end{myitemize}
The rate-distortion operating points used in this work are defined by combinations of temporal aggregation intervals and JPEG~2000 target bitrates of 2, 1.5, 1, 0.5, 0.25, and 0.1 bits per pixel (bpp), together with the lossless mode. The temporal aggregation intervals are selected according to the event density of each dataset being coded: 5, 10, 20, 50, and 100~ms for the relatively dense ECD dataset; 5, 10, 20, and 50~ms for MVSEC dataset; and 50, 100, and 150~ms for the considerably sparser Gen1 dataset, for which shorter intervals would produce histogram frames containing too few events to achieve efficient compression.

When JPEG 2000 operates in lossless mode, the histogram frames are perfectly reconstructed, preserving the event counts at every pixel. Consequently, any variation in downstream task performance can be attributed entirely to the temporal aggregation process. In contrast, under lossy compression, errors may be introduced into the reconstructed histogram frames, altering the reconstructed event counts and their spatial distribution, which may further degrade downstream task performance.
\section{Point Cloud-Based Event Stream Compression Pipeline}\label{sec:codec-gpcc}

The second compression pipeline considered in this work was built around a fundamentally different codec family: the MPEG geometry-based PC codec, referred to as G-PCC~\cite{GPCC}. In contrast to the aggregation-based event codec described in Section~\ref{sec:codec}, which first converts the event stream into a sequence of histogram frames, the point cloud-based event compression pipeline directly represents and encodes events in the native $(x,y,t)$ domain, without requiring an intermediate frame-based representation.

\begin{figure*}[t]\centering
\includegraphics[width=0.9\textwidth]{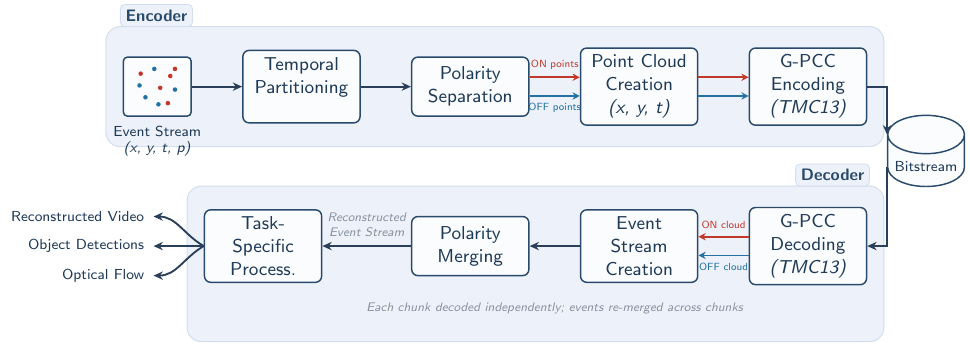}
\caption{Point cloud-based event stream compression pipeline using G-PCC.}\label{fig:gpcc}
\end{figure*}

Figure~\ref{fig:gpcc} illustrates the proposed G-PCC-based event compression pipeline. The encoder first partitions the event stream into fixed-duration temporal chunks. Within each chunk, events are separated by polarity, yielding two 3D point clouds, one for each (chunk, polarity) pair, where each event is represented by its $(x,y,t)$ coordinates. Each 3D point cloud is then independently compressed using G-PCC. The decoder performs the inverse process: each compressed bitstream is G-PCC decoded into a reconstructed 3D point cloud, the corresponding polarity is recovered from the associated stream, and the reconstructed events from all chunks and polarities are merged and sorted by timestamp to form the final reconstructed event stream.
\subsection{Codec Architecture and Walkthrough}\label{ssec:gpcc_arch}

In the proposed PC-based event compression solution, the following operations are performed:

\boldstep{Temporal Partitioning} The event stream is first partitioned into fixed-duration temporal chunks. Chunks are non-overlapping and temporally aligned with the first event timestamp. This partitioning bounds the temporal extent of each point cloud, controls the end-to-end system latency, and keeps the encoder memory footprint independent of the total sequence length. The chunk duration is a system-level parameter that trades off compression efficiency, latency, and encoder memory; in this work, it is set to $100$~ms, as this value was found experimentally to keep each point cloud small enough for efficient octree coding while maintaining acceptable latency.

\boldstep{Polarity Separation and Point Cloud Creation} For each temporal chunk, events are separated by polarity into two streams. Each stream is mapped to a 3D point cloud with coordinates $(x, y, z)$, where $x$ and $y$ correspond to the original spatial coordinates of each event and $z$ is derived from its timestamp. To create the PC representation, the three coordinates are mapped onto a common 10-bit integer grid $[0, 1023]$ (that is, $1024$ levels, $0$ to $2^{10}-1$) using fixed, data-independent ranges. The spatial coordinates are scaled from the sensor's native pixel grid with a single common factor $S = \max(W, H) - 1$, where $W$ and $H$ denote the sensor width and height, so that the sensor aspect ratio is preserved and one pixel corresponds to the same grid step along both spatial axes:
\begin{equation}
q_x = \operatorname{round}\!\left(\frac{x \cdot 1023}{S}\right), \qquad
q_y = \operatorname{round}\!\left(\frac{y \cdot 1023}{S}\right).
\end{equation}
With a shared scale factor, only the longer spatial axis spans the full grid, while the shorter one occupies a proportionally smaller sub-range. The temporal coordinate is obtained by mapping the within-chunk timestamp offset $t - t_{\text{chunk}}$ over the full chunk duration $\Delta$:
\begin{equation}
q_z = \operatorname{round}\!\left(\frac{(t - t_{\text{chunk}}) \cdot 1023}{\Delta}\right).
\end{equation}
Since $W$, $H$, $\Delta$ and the grid size are constants known to both the encoder and the decoder, no side information is required to invert these mappings. The resulting coordinate values are integers and the point clouds are stored in binary PLY format using \texttt{float32} coordinate fields, which preserve these integer-valued coordinates without precision loss. The motivation for mapping all three coordinates onto a common grid is that G-PCC performs octree subdivision jointly along the three coordinate axes. If one axis dominates the numeric range, for example raw timestamps spanning $\sim\!10^5$~$\mu$s per chunk compared with spatial dimensions of a few hundred pixels, the octree partitioning becomes unbalanced, with a disproportionate number of subdivisions devoted to the dominant axis and limited refinement of the remaining ones. Mapping the three coordinates onto a common range balances the spatial and temporal scales of the point cloud, allowing the octree to allocate subdivisions more effectively across all dimensions and providing comparable quantization resolution along the spatial and temporal axes. Note that polarity is not encoded as a point cloud attribute; instead, it is implicitly represented by the identity of the stream, i.e., positive (ON) or negative (OFF) polarity. Therefore, each (chunk, polarity) pair produces an independent PC containing only geometry coordinates and no additional attributes.

\boldstep{G-PCC Encoding and Decoding} Each PC created in the previous step is independently compressed using the MPEG G-PCC reference software (TMC13 v23.0-rc2)~\cite{gpcc:tmc13}. The encoder is configured in octree geometry coding mode with its default settings; the only parameter varied across operating points is the \texttt{positionQuantizationScale} parameter, which controls the geometry quantization scale and determines the tradeoff between bitrate and reconstructed geometric distortion. Duplicate-point merging is enabled (\texttt{mergeDuplicatedPoints}$=1$). As a result, when multiple input events are mapped to the same quantized 3D voxel, they are represented by a single point in the decoded PC, which may therefore contain fewer points than the input. This effect becomes more pronounced at coarser quantization levels, where a larger number of events are mapped to the same quantized position. Because timestamps are quantized onto a fixed temporal grid within each chunk, spatially co-located events (i.e., occurring at the same $(x, y)$ position) that fall in the same temporal grid cell merge into a single point; consequently, the reconstructed event count is not necessarily identical to the input count, although the difference is typically small at high bitrates. The bitrate is reported in bits per event, computed with respect to the original (pre-merge) event count to ensure comparability across operating points and codecs. At the decoder, each compressed bitstream is G-PCC decoded into a reconstructed 3D point cloud.

\boldstep{Event Stream Creation and Polarity Merging} The reconstructed point clouds are converted back into event streams by inverting the fixed mappings applied during the Point Cloud Creation step. Specifically, the decoded coordinates $(\hat{q}_x, \hat{q}_y, \hat{q}_z)$ are mapped back to the sensor's native pixel grid and to an absolute event timestamp as:
\begin{align*}
\hat{x} = \operatorname{round}\!\left(\frac{\hat{q}_x \cdot S}{1023}\right),
\hat{y} = \operatorname{round}\!\left(\frac{\hat{q}_y \cdot S}{1023}\right),
\hat{t} = t_{\text{chunk}} + \frac{\hat{q}_z \cdot \Delta}{1023}.
\end{align*}
Since the chunk duration, the sensor dimensions and the grid size are constants known to both encoder and decoder, this inversion requires no per-chunk side information; only the absolute start time of the sequence must be signaled, from which the start time of every chunk is derived. Each reconstructed event is assigned the polarity of the stream from which it was decoded. Finally, the events from all temporal chunks and both polarities are concatenated and globally sorted by timestamp to produce the reconstructed event stream in the usual $(x, y, t, p)$ format.

\subsection{Codec Operating Points}\label{ssec:gpcc_points}

The rate-distortion operating points of the PC-based event codec are controlled by the geometry quantization scale parameter \texttt{positionQuantizationScale}, denoted $q$ for conciseness. This parameter defines the quantization scale applied to the input point cloud coordinates $(x, y, z)$ before octree construction; $q = 1.0$ corresponds to lossless geometry coding while lower values reduce the coordinates precision during geometry quantization, leading to lower bitrate at the expense of increased geometric distortion. The operating points used in this work are $q \in \{1.0, 0.375, 0.3, 0.2, 0.125, 0.0625, 0.03125\}$.

Unlike the aggregation-based pipeline, which discretizes time into fixed-duration aggregation intervals, the PC-based pipeline processes each event timestamp individually. Because the three point cloud coordinates share a common grid and scale, G-PCC performs geometry quantization jointly across the spatial and temporal dimensions, equally affecting the $(x, y, t)$ coordinates, rather than introducing an explicit temporal aggregation step as in the aggregation-based pipeline. Consequently, the PC-based pipeline preserves the fine-grained temporal representation than the millisecond-order aggregation windows used by the JPEG 2000-based pipeline. This difference in how the two codecs treat the temporal component makes point cloud coding a valuable alternative for assessing whether a distortion metric remains sensitive to event stream distortions independently of the underlying codec representation.

\section{Task-Driven Evaluation Framework}\label{sec:task-eval}

Evaluating event-based compression requires assessing not only traditional signal-level distortion metrics but also the impact of compression on downstream task performance. To this end, a reference-relative (RR) evaluation protocol is adopted, in which the downstream task is first run on the uncompressed event stream to obtain a reference output, and then run again on the decoded event stream (lossy) to obtain a test output. Task performance is then measured by comparing the test output against the reference output, rather than against an external ground truth. This protocol isolates the effect of compression from the intrinsic errors of the downstream model, allowing performance changes to be attributed exclusively to the compression process. The RR protocol is used for all downstream tasks considered in this work (see Section~\ref{ssec:tasks}); only the task-specific evaluation metric differs.

\subsection{Selected Tasks}\label{ssec:tasks}

To evaluate the impact of event compression on downstream task performance, three representative tasks are now considered (a fourth one is considered in Section~\ref{ssec:async}, covering both human visual interpretation and machine vision. Each task is evaluated using a well-established model and a widely adopted benchmark dataset.
\begin{myitemize}
    \item \textbf{Video reconstruction:} Evaluates the impact of compression on visual reconstruction quality. To this end, HyperE2VID~\cite{Ercan:2024} is used as the reconstruction model. HyperE2VID reconstructs intensity videos from event streams using a recurrent neural network with hypernetworks. Evaluation is performed on the Event Camera Dataset (ECD)~\cite{Mueggler:2017}, which comprises seven sequences captured with a DAVIS240C sensor with 240$\times$180 pixel resolution. Reconstruction quality is measured using the Peak Signal-to-Noise Ratio (PSNR) and the Structural Similarity Index (SSIM) between the test and reference reconstructions.
    \item \textbf{Object detection:} Evaluates the impact of compression on time-critical detection in real-time applications. To this end, RVT~\cite{GehrigM:2023} is used as the object detection model. RVT performs event-based object detection using hierarchical vision transformers with recurrent memory. Evaluation is performed on three sequences from the Gen1 dataset~\cite{Tournemire:2020}, containing annotated bounding boxes for cars and pedestrians; sequences were captured with a PROPHESEE GEN1 sensor with 304$\times$240 pixel resolution. Detection performance is measured using the mean Average Precision (mAP) metric, following the COCO evaluation procedure~\cite{Lin:2014}. The mAP metric is computed by comparing the test and reference detections and averaged across all sequences.
    \item \textbf{Optical flow estimation:} Assesses the impact of compression on motion estimation accuracy. To this end, E-RAFT~\cite{GehrigM:2021} is used as the optical flow estimation model. E-RAFT estimates dense optical flow from voxel-grid representations via correlation volumes. Evaluation is performed on two driving sequences from the MVSEC dataset~\cite{ZhuAZ:2018}, captured using a DAVIS346 sensor with a 346$\times$260 pixel resolution. Performance is measured using the End-Point Error (EPE), defined as the mean Euclidean distance between the test and reference flow fields.
\end{myitemize}

For a comprehensive performance analysis, the two compression pipelines introduced in Sections~\ref{sec:codec} and~\ref{sec:codec-gpcc} are evaluated independently over their respective operating points, listed in Sections~\ref{ssec:codec_points} and~\ref{ssec:gpcc_points}, and and their impact on the three downstream tasks is assessed using the same RR protocol.

\begin{figure*}[!ht]
    \centering
    \begin{subfigure}[t]{0.32\textwidth}
        \includegraphics[width=\textwidth]{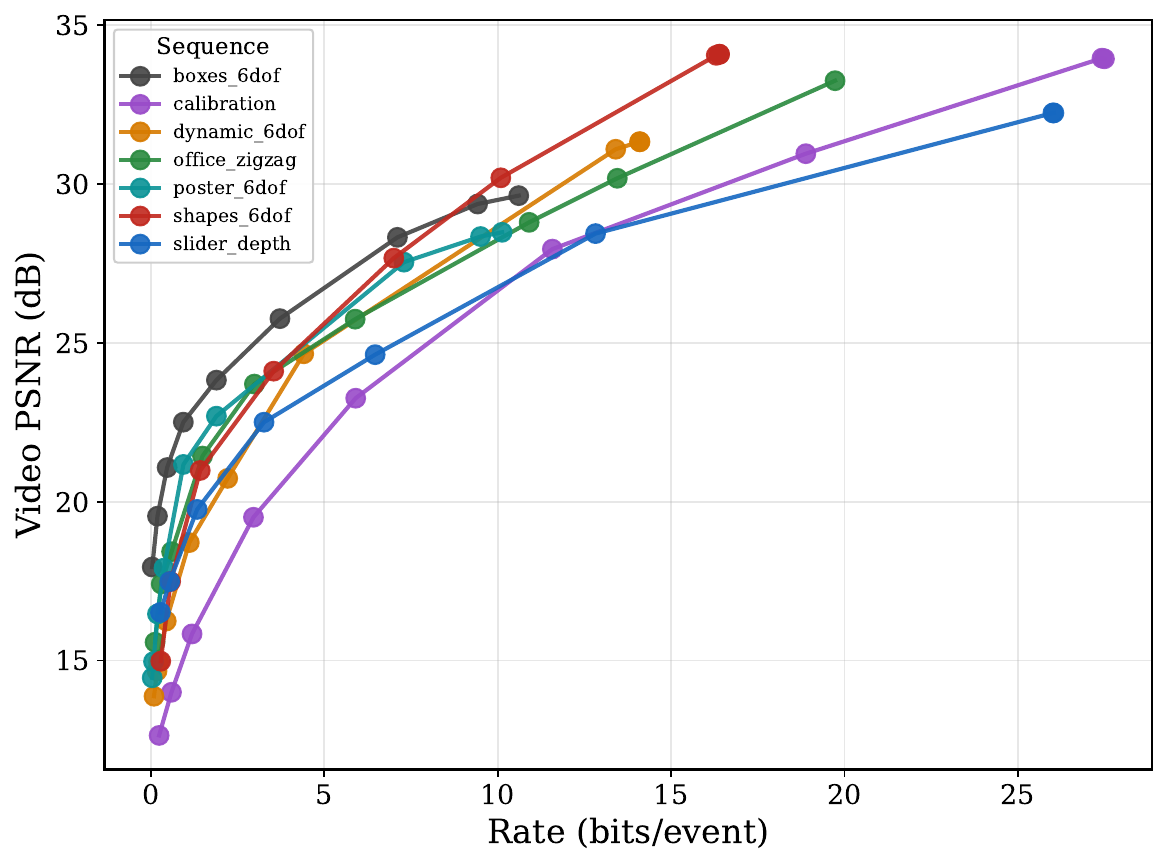}
        \subcaption{Video reconstruction (HyperE2VID, ECD).}
    \end{subfigure}\hfill
    \begin{subfigure}[t]{0.32\textwidth}
        \includegraphics[width=\textwidth]{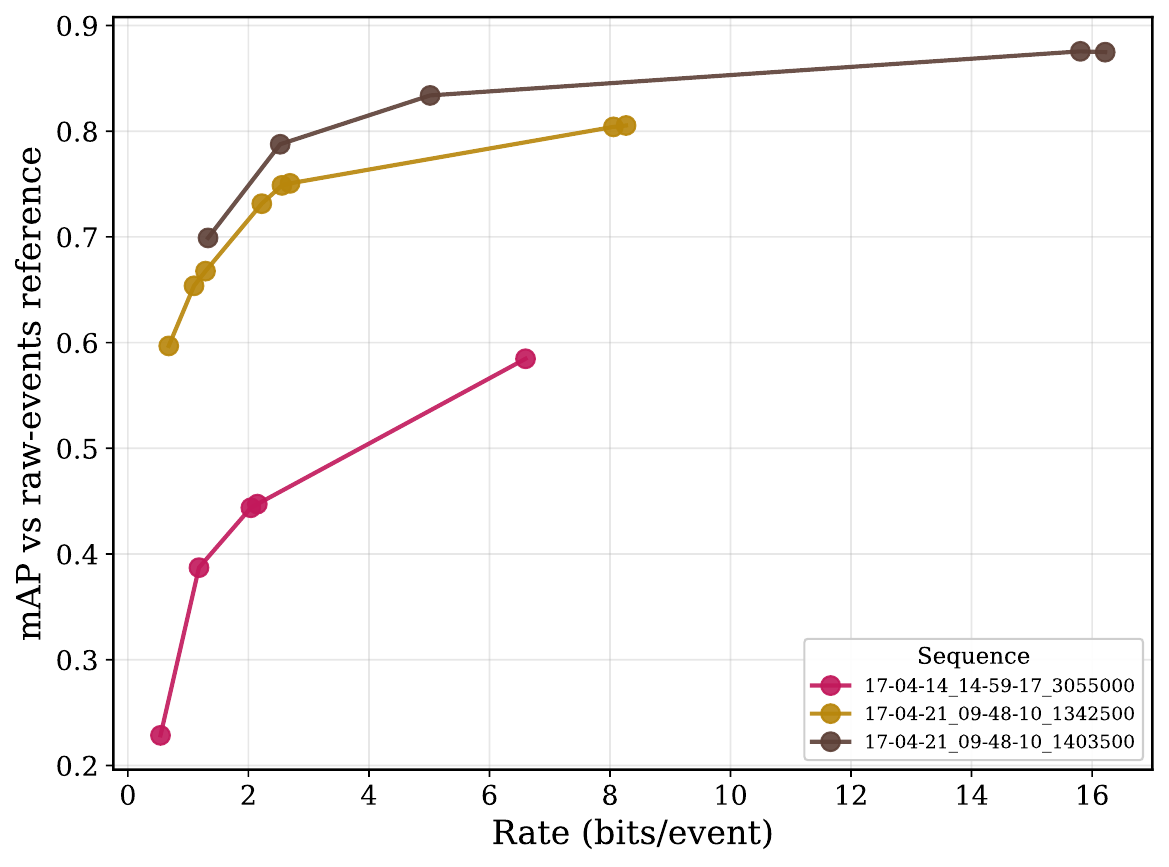}
        \subcaption{Object detection (RVT, Gen1).}
    \end{subfigure}\hfill
    \begin{subfigure}[t]{0.32\textwidth}
        \includegraphics[width=\textwidth]{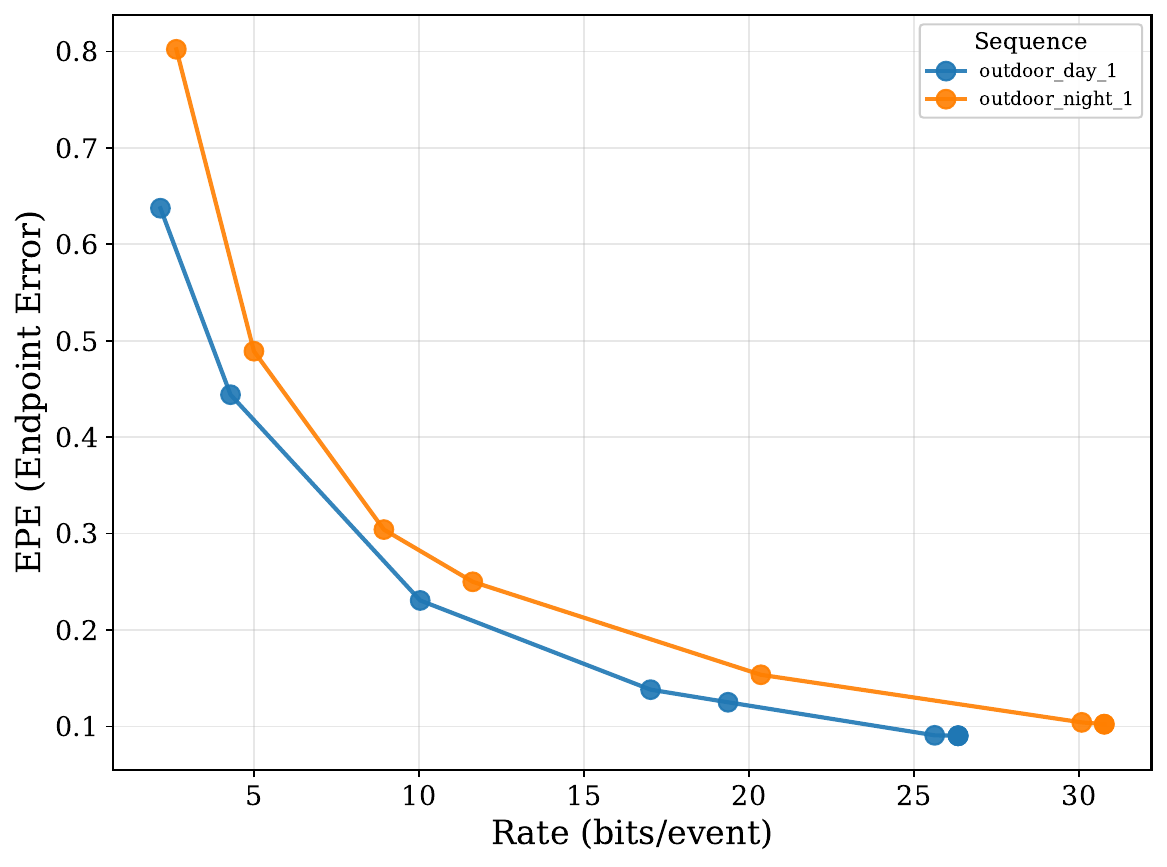}
        \subcaption{Optical flow estimation (E-RAFT, MVSEC).}
    \end{subfigure}
    \caption{Rate-utility curves for the aggregation-based codec (JPEG 2000-based) across the three downstream tasks, evaluated over the temporal aggregation intervals and operating points defined in Section~\ref{ssec:codec_points} under the RR protocol. Each curve is obtained only with the operating points of the convex hull.}
    \label{fig:rd-jp2k}
\end{figure*}

\begin{figure*}[!ht]
    \centering
    \begin{subfigure}[t]{0.32\textwidth}
        \includegraphics[width=\textwidth]{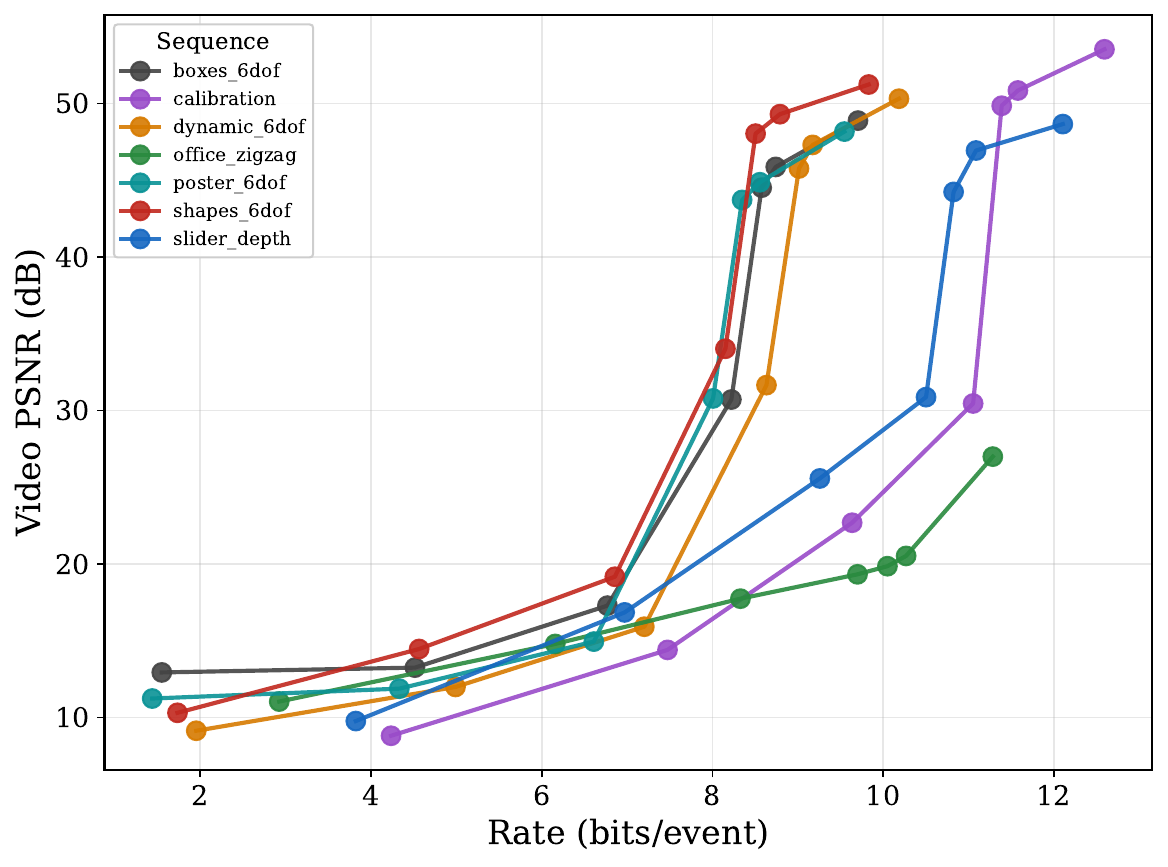}
        \subcaption{Video reconstruction (HyperE2VID, ECD).}
    \end{subfigure}\hfill
    \begin{subfigure}[t]{0.32\textwidth}
        \includegraphics[width=0.95\textwidth]{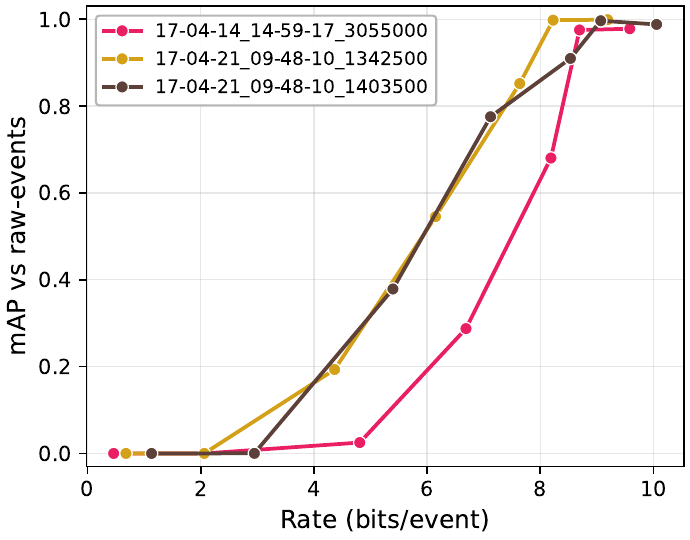}
        \subcaption{Object detection (RVT, Gen1).}
    \end{subfigure}\hfill
    \begin{subfigure}[t]{0.32\textwidth}
        \includegraphics[width=\textwidth]{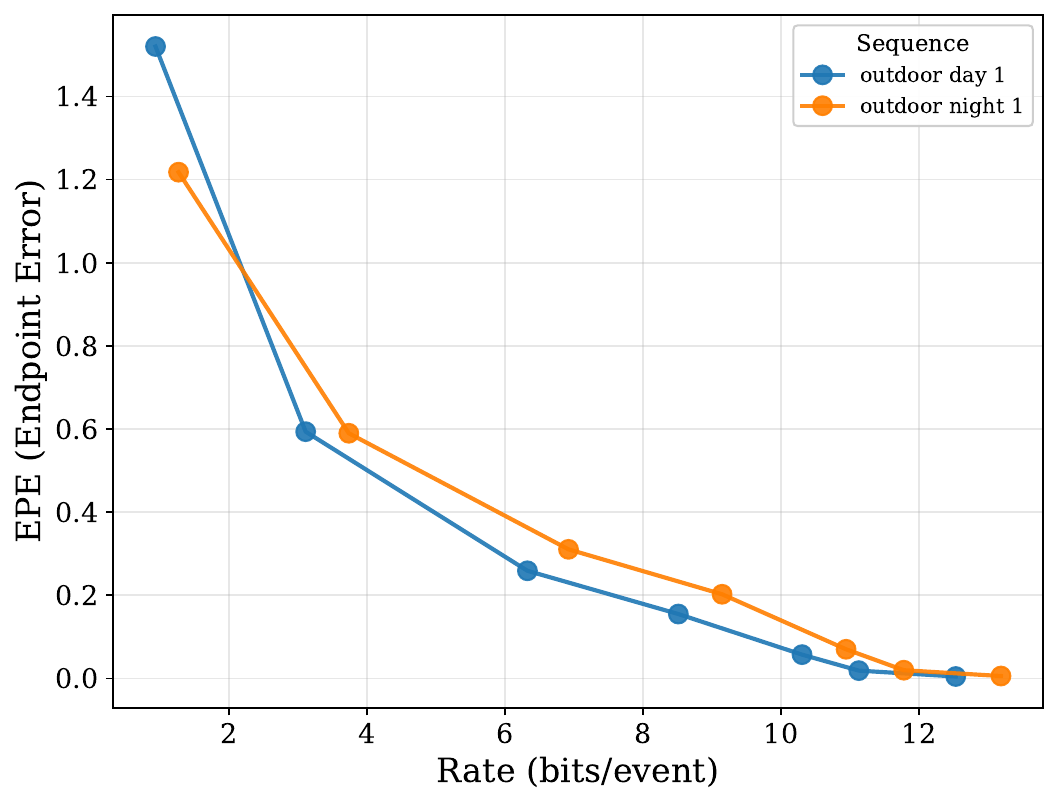}
        \subcaption{Optical flow estimation (E-RAFT, MVSEC).}
    \end{subfigure}
    \caption{Rate-utility curves for the point cloud-based codec (G-PCC-based) across the three downstream tasks, evaluated over the operating points defined in Section~\ref{ssec:gpcc_points} under the RR protocol. Each curve corresponds to one sequence.}
    \label{fig:rd-gpcc}
\end{figure*}

\subsection{Impact Assessment of Event Codecs}\label{ssec:codec-assessment}

The impact of event compression on the performance of each task is quantified using the RR protocol described in the previous Section~\ref{sec:task-eval}. For each task and each codec, performance is evaluated at every operating point by comparing the task output obtained from the reconstructed event stream against the corresponding reference output. The resulting rate-utility curves, which relate coding rate to downstream task performance, are shown in Fig.~\ref{fig:rd-jp2k} for the aggregation-based codec and in Fig.~\ref{fig:rd-gpcc} for the PC-based codec. As shown in these figures, the downstream task performance relative to the uncompressed event stream reference degrades as the rate decreases across all tasks and both codecs, at a pace that depends on the sequence characteristics and codec configuration.

In the lossless operating mode of both JPEG2000 and G-PCC, the coding engines themselves introduce no compression loss. While JPEG~2000 operating in reversible mode reconstructs the histogram (count) frames bit-exactly, G-PCC with $q = 1.0$ reconstructs the PC geometry losslessly. Consequently, any residual degradation in downstream task performance at these operating points result from the pre-processing steps (which include some type of quantization) rather than from the coding process itself. For the aggregation-based pipeline, this residual degradation is caused by the temporal aggregation, which discards original event timestamps. For the PC-based pipeline, it results from mapping event timestamps onto a fixed-resolution temporal grid prior to coding, together with the associated merging of events that become coincident after temporal quantization; note that, since the spatial coordinates are recovered exactly at the lossless operating point of the point-cloud codec, the residual distortion originates from the temporal coordinate representation.

\subsection{Impact Assessment of Image Codecs}\label{ssec:codec_impact}

The aggregation-based codec proposed in Section~\ref{sec:codec} employs JPEG~2000 as its underlying image codec. Although newer conventional image codecs, such as JPEG XL and AVIF, generally provide higher compression efficiency on natural images, it is unclear whether this advantage extends to the sparse event histogram frames generated by the temporal aggregation process. In this context, this section presents a preliminary comparison between JPEG~2000, a wavelet-based image codec, and the block transform-based codecs JPEG XL (VarDCT mode) and AVIF. The goal of this comparison is not to benchmark image codecs exhaustively, but rather to compare two representative and very recent image coding approaches: wavelet-based and block transform-based coding. The comparison is performed at two levels: first, by analyzing the effect of image coding on the reconstructed histogram frames, and then by evaluating its impact on downstream task performance.

\subsubsection{Impact on Image Codec Output}
\begin{figure}[!htb]
    \centering
    \begin{minipage}[b]{0.48\linewidth}
        \centering
        \includegraphics[width=\textwidth]{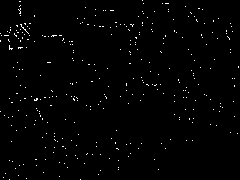}
        \subcaption{Original}
    \end{minipage}
    \hfill
    \begin{minipage}[b]{0.48\linewidth}
        \centering
        \includegraphics[width=\textwidth]{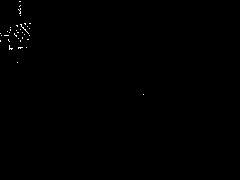}
        \subcaption{AVIF}
    \end{minipage}

    \vspace{0.5em}

    \begin{minipage}[b]{0.48\linewidth}
        \centering
        \includegraphics[width=\textwidth]{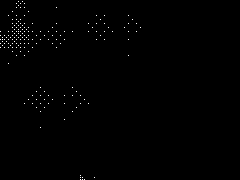}
        \subcaption{JPEG XL}
    \end{minipage}
    \hfill
    \begin{minipage}[b]{0.48\linewidth}
        \centering
        \includegraphics[width=\textwidth]{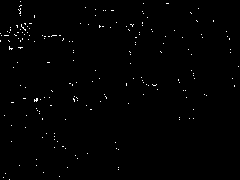}
        \subcaption{JPEG 2000}
    \end{minipage}

    \caption{Illustration of the impact of three different image coding schemes on a sparse event count decoded histogram (ECD boxes\_6dof, 
    frame 0, negative polarity, 98.9\% zeros, 483 events). JPEG 2000 operates at 0.2\,bpp, while JPEG XL and AVIF operate at their minimum distortion level.}
    \label{fig:codec-comparison}
\end{figure}

A preliminary experiment was conducted by compressing the histograms (count) frames, produced by the temporal aggregation stage (Section~\ref{sec:codec}), using JPEG~2000, JPEG XL and AVIF. JPEG XL\footnote{libjxl v0.12.0 (commit 53042ec5), \url{https://github.com/libjxl/libjxl}.} and AVIF\footnote{libavif 1.3.0, \url{https://github.com/AOMediaCodec/libavif}.} were configured in lossy mode at their highest quality setting (QP = 99), while JPEG~2000\footnote{Kakadu v8.4.1, \url{https://kakadusoftware.com}.} operated at 0.2 bpp. As shown in Fig.~\ref{fig:codec-comparison}, the AVIF-decoded histogram frame retained only 19\% of the original events, while JPEG XL retained 37\% and introduced additional false (phantom) events were not present in the original histogram frame, despite operating at its \textit{highest achievable quality}. In contrast, JPEG~2000 preserved 80\% of the original events at 0.2 bpp and, at higher bitrates, reconstructs the original histogram frames accurately. These observations suggest that block transform-based compression methods may not be well suited to the sparse structure of event histogram frames. In natural images, neighboring pixels tend to exhibit similar intensities, while event histogram frames are dominated by zero-valued pixels, whose locations carry semantic information (the absence of events). Applying block transform-based compression to the histogram frames tends to spread the energy of isolated non-zero coefficients into neighboring regions, through quantization and inverse transform, thus creating spurious non-zero values and altering the sparse spatial structure that downstream event-based processing relies upon.

\subsubsection{Impact on Task Performance}\label{sssec:rd-jp2k-jxl}

\begin{figure*}[!t]
    \centering
    \includegraphics[width=0.94\textwidth]{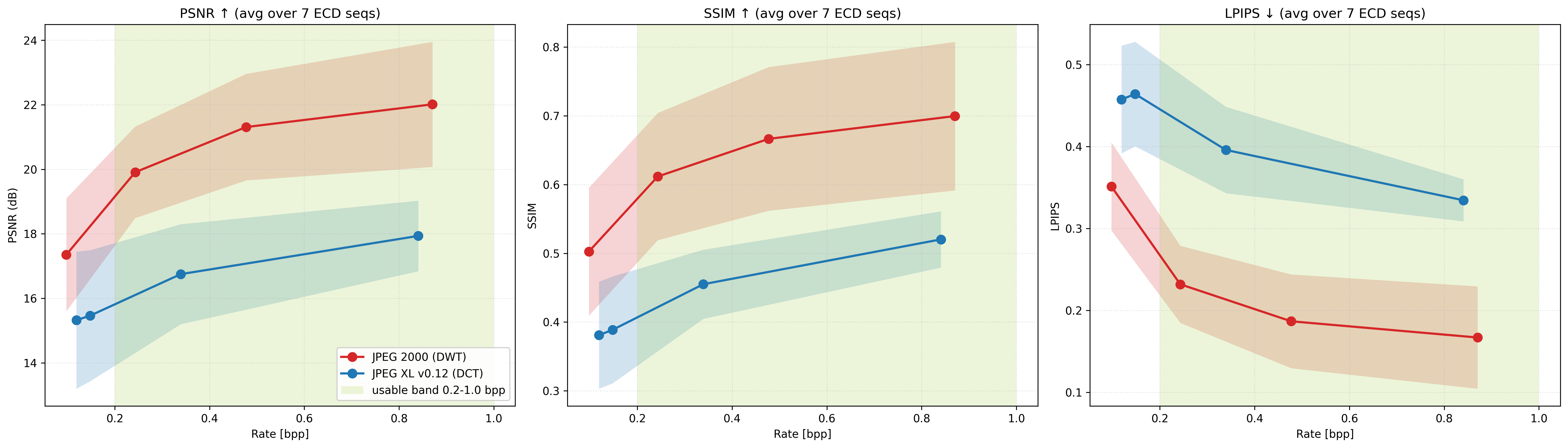}
    \caption{Rate-distortion curves for the video reconstruction task (HyperE2VID) on the ECD dataset, averaged over the seven ECD sequences. Shaded regions indicate $\pm 1$ standard deviation across sequences.}
    \label{fig:rd-jp2k-jxl}
    \vspace{-10pt}
\end{figure*}

\begin{figure}[!htb]
    \centering
    \begin{subfigure}{\linewidth}
        \centering
        \includegraphics[width=\textwidth]{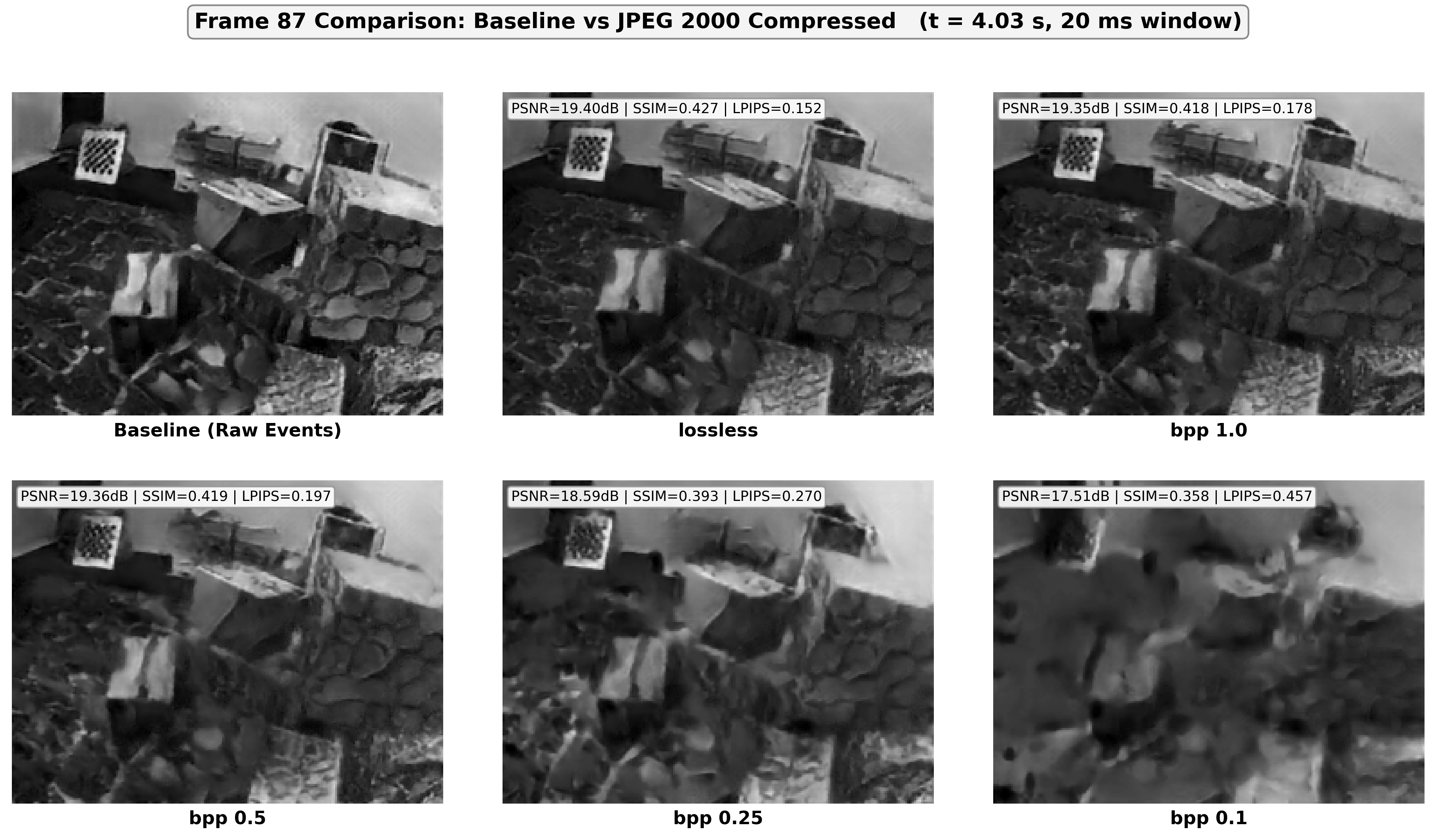}
        \subcaption{JPEG~2000 frame reconstructions.}
    \end{subfigure}
    
    \vspace{0.5em}
    
    \begin{subfigure}{\linewidth}
        \centering
        \includegraphics[width=\textwidth]{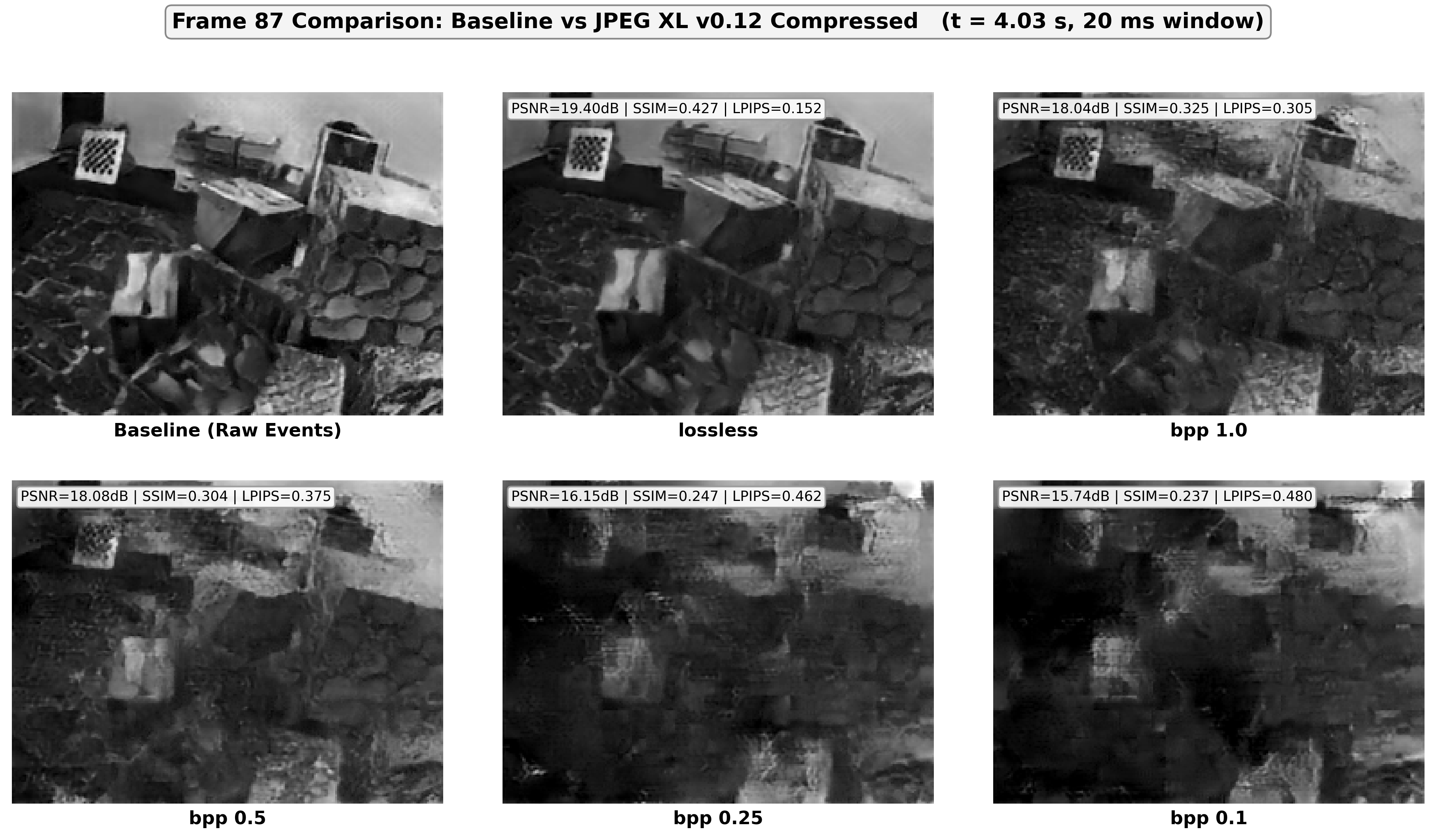}
        \subcaption{JPEG~XL frame reconstructions.}
    \end{subfigure}
    \caption{HyperE2VID reconstruction at four lossy bitrates and lossless coding mode on ECD boxes\_6dof at $t = 4.03$~s (20~ms aggregation interval). All images correspond to the same time instant with identical grayscale scaling for direct codecs comparison. }
    \label{fig:visual-jp2k-jxl}
\end{figure}

While Fig.~\ref{fig:codec-comparison} illustrates the effect of different image codecs on a single event histogram frame, a quantitative evaluation is needed to determine whether these differences translates into differences in downstream task performance. For this purpose, the complete aggregation-based event stream compression pipeline shown in Fig.~\ref{fig:agg-codec-pipeline} was evaluated for the video reconstruction task, using both JPEG~2000 and JPEG~XL. Each codec was used to compress the event histogram frames, which were then converted back into an event stream and processed by the HyperE2VID model to reconstruct intensity videos. Reconstruction quality was then evaluated by comparing the reconstructed videos against those obtained from the original (uncompressed) event streams (baseline) using PSNR, SSIM and LPIPS.
The evaluation was performed on all seven ECD sequences using a 20~ms temporal aggregation interval and 8-bit histogram frames. JPEG~2000 was configured as described in Section~\ref{ssec:codec_arch}, while JPEG~XL was operated in its lossy VarDCT mode\footnote{libjxl v0.12.0, with perceptual optimizations disabled (\texttt{-disable\_perceptual\_optimizations} flag).}. Figure~\ref{fig:rd-jp2k-jxl} shows the resulting rate-distortion curves, averaged over the seven ECD sequences, for PSNR, SSIM and LPIPS quality metrics. As it can be observed, at the same bitrate, JPEG~2000 outperforms JPEG~XL by up to +4.5~dB in PSNR, with largest differences observed in the practically relevant bitrate range between 0.25--1.0~bpp. Figure~\ref{fig:visual-jp2k-jxl} provides representative visual examples, showing that JPEG~2000 reconstructions preserves the scene structure and degrade gracefully as bitrate decreases, while JPEG~XL reconstructions show progressive loss of structural detail.

Although JPEG~XL generally provides higher compression efficiency than JPEG~2000 on natural images, this behavior is reversed for sparse event histogram frames. a reversal that stems from the DCT at the core of JPEG~XL. Event histogram frames consist predominantly of zero-valued pixels, where the absence of events is itself semantically meaningful. During transform coding, isolated non-zero pixel values contribute to multiple transform coefficients, and coefficient quantization spreads this energy over neighboring pixels after inverse transform. As a result, pixels that were originally zero may acquire small non-zero values in the decoded histogram frame. This effect is substantially less pronounced in JPEG~2000, whose wavelet transform preserves sparse structures more effectively.

After inverse aggregation, these spurious non-zero pixels generate phantom events that were not present in the original event stream, introducing false activity that propagates to downstream processing, as observed in Fig.~\ref{fig:visual-jp2k-jxl}. This behavior can be measured directly from the decoded histogram frames. For each codec setting, the count ratio is defined as $\sum_{x,y} \hat{I}(x,y) / \sum_{x,y} I(x,y)$, where $I$ and $\hat{I}$ denote the original and decoded histogram frames, respectively; a value of $1.0$ indicates preservation of the total event count, while values above $1.0$ indicate generation of phantom events. Across the seven ECD sequences, the count ratio for JPEG~XL increases from $1.00$ at lossless coding to between $1.13$ and $1.26$ at lossy coding, whereas JPEG~2000 stays close to $1.0$ across all operating points. By preserving the sparse structure of event histogram frames and avoiding the generation of phantom events, JPEG~2000 provides substantially better downstream performance and, for that reason, is adopted as the image codec in the aggregation-based compression pipeline proposed in Section~\ref{sec:codec}.

\section{Event Stream Distortion Metrics}\label{sec:metrics}

Several objective metrics have been proposed in previous works to evaluate reconstructed event stream fidelity, notably Asynchronous Spatiotemporal Spike Metric (ASTSM)~\cite{Li:2023,stumpp2024flow}, PSNR Event-to-Event (PSNR E2E) \cite{seleem2026deep}, Spatial Fidelity ($VQ_s$) and Temporal Fidelity ($VQ_t$) \cite{adhuran2025lossy}, Frame PSNR and Frame MSE \cite{banerjee2021lossy}. These metrics capture complementary aspects of spatio-temporal reconstruction quality and are reviewed in Section~\ref{sec:background}. However, they generally measure overall reconstruction fidelity without explicitly distinguishing between different types of compression-induced errors, such as event loss and spurious event generation.

\subsection{Proposed Classification-based Metrics}\label{ssec:classifmetrics}

The example illustration of Fig.~\ref{fig:codec-comparison} reveals two distinct compression-induced error types: \textit{event loss}, where original events are missing due to quantization, and \textit{event hallucination}, where spurious events appear at locations that were inactive in the original stream. Traditional signal-level metrics like PSNR combine these different distortion sources into a single score, making it difficult to determine their individual impact on downstream task performance. To distinguish these different distortion sources, a classification-based approach is adopted.

The proposed metrics require only the original and reconstructed event streams, making them applicable to any codec architecture, including temporal aggregation-based, PC-based, or learned compression-based. To compute the metrics, both streams are aggregated into binary frames over fixed-duration temporal intervals, merging both polarities into a single frame; unlike the histogram frames created by the aggregation-based codec in Section~\ref{sec:codec}, where each pixel stores the number of events, here each pixel is assigned a binary label indicating whether at least one event of either polarity occurred within the corresponding temporal interval (positive class) or not (negative class). This binary formulation focuses the evaluation on the preservation of event activity locations, which represent a fundamental component of event based visual information, rather than on the exact number of events at each location. Because active-event pixels are typically much sparser than inactive pixels, this formulation yields a highly imbalanced binary classification problem, motivating the use of metrics commonly adopted in imbalanced classification and anomaly detection \cite{Powers2011,Chicco2020,Cohen1960}. A pixel-wise comparison between the original and reconstructed binary frames yields a confusion matrix containing: True Positives (TP), where both frames contain events; False Negatives (FN), where original events are missing after reconstruction; False Positives (FP), where spurious events are introduced; and True Negatives (TN), where both frames are empty.

The temporal interval used for metric computation (15~ms for the ECD, Gen1, and MVSEC datasets in this work) is intentionally decoupled from the codec's internal temporal aggregation setting. This ensures that the metrics remain codec-agnostic and enable consistent comparison across different compression architectures. Based on the accumulated confusion matrix, five metrics that capture complementary aspects of compression fidelity are considered:
\begin{myitemize}
    \item \textbf{Recall} measures event survival rate by computing the fraction of original event locations retained after compression: $\text{Recall} = \text{TP}/(\text{TP} + \text{FN})$. This metric isolates event loss, which becomes dominant under aggressive quantization or temporal aggregation.
    
    \item \textbf{Intersection over Union (IoU)} measures the overlap between original and reconstructed event sets, penalizing both missing and spurious events: $\text{IoU} = \text{TP}/(\text{TP} + \text{FP} + \text{FN})$. This metric directly reflects preservation of spatial event structure, including motion contours, object boundaries, and temporal bursts that downstream models rely upon.
    
    \item \textbf{Markedness} evaluates the reliability of the reconstructed event labels: $M = \text{PPV} + \text{NPV} - 1$ where $\text{PPV} = \text{TP}/(\text{TP} + \text{FP})$ and $\text{NPV} = \text{TN}/(\text{TN} + \text{FN})$. Since false positives reduce positive predictive value (PPV), markedness is particularly sensitive to compression artifacts that introduce spurious events in inactive (empty) regions.
    
    \item \textbf{Matthews Correlation Coefficient (MCC)} provides a balanced single-value summary of event stream fidelity that remains meaningful under extreme class imbalance:
    \begin{equation}
    \text{MCC} = \frac{\text{TP} \times \text{TN} - \text{FP} \times \text{FN}}{\sqrt{(\text{TP}+\text{FP})(\text{TP}+\text{FN})(\text{TN}+\text{FP})(\text{TN}+\text{FN})}}
    \end{equation}
    Unlike accuracy, MCC cannot be inflated by simply preserving the dominant empty class.
    
    \item \textbf{Cohen's Kappa} evaluates whether the codec performs better than a trivial baseline that predicts no events everywhere: $\kappa = (p_o - p_e)/(1 - p_e)$, where $p_o$ is observed accuracy (fraction of matching labels) and $p_e$ is expected agreement by chance based on class label frequencies. A codec achieving high accuracy by outputting zeros everywhere would score $\kappa \approx 0$, exposing its failure to preserve the rare but semantically critical event pixels.
\end{myitemize}

Together, these five metrics form a complementary evaluation framework: Recall isolates event loss, IoU captures structural fidelity, Markedness detects hallucination artifacts, while MCC and Cohen's $\kappa$ provide imbalance-robust summary measures. This evaluation framework enables codec comparison based on event-level fidelity without requiring computationally expensive downstream task evaluation.

\subsection{Metric-to-Task Correlation Assessment}\label{ssec:correlation}
Each event stream distortion metric is evaluated by measuring its ability to predict compression-induced task degradation under the RR protocol described in Section~\ref{ssec:codec-assessment}. For each downstream task, a correlation pool is first assembled, containing the paired distortion metric values $X$ and task performance values $Y$ measured at selected lossy operating points, collected over all sequences of the corresponding dataset. For the aggregation-based codec, whose operating points combine a temporal aggregation interval and a target bitrate, only the rate-utility Pareto-optimal points of each sequence are included, consistent with the convex hulls shown in Fig.~\ref{fig:rd-jp2k}, since dominated operating points would never be selected in practice by a rate-distortion optimized encoder. The distortion metric values in each pool are mapped to predicted task performance scores $\hat{Y}$ using a cubic polynomial regression, following ITU-T Rec.~P.1401:
\begin{equation}
\hat{Y} = a_0 + a_1 X + a_2 X^2 + a_3 X^3
\end{equation}
where the coefficients are estimated via least-squares fitting against the observed task performance values $Y$. This mapping is necessary because the scale and curvature of an event stream metric differs of the task scoring metric; mapping before correlation lets each metric be assessed on its predictive power rather than on its scale. Two complementary measures are then used to quantify prediction accuracy for each downstream task, both computed between the fitted predictions $\hat{Y}$ and the observed task-specific values $Y$: i) the Pearson Linear Correlation Coefficient (PLCC), which evaluates their linear correlation, and ii) the Spearman Rank-Order Correlation Coefficient (SROCC), which evaluates monotonic relationships based on rank ordering. Note that, since SROCC depends only on rank ordering, it is invariant under monotonic mappings; whenever the fitted cubic is monotonic over the data range, as observed for the proposed metrics, the SROCC computed on the fitted values coincides with the SROCC computed directly between $X$ and $Y$. Event stream metrics achieving SROCC $\geq 0.80$ can be considered reliable predictors of task performance degradation. To establish whether predictive power depends on the codec, correlations are reported separately for the aggregation-based codec (JPEG~2000, Section~\ref{sec:codec}) and the PC-based codec (G-PCC, Section~\ref{sec:codec-gpcc}), and additionally over the pooled operating points of both codecs; in the pooled case, the selected points of the two codecs are concatenated into a single pool, over which one mapping is applied and the two correlation measures are computed, rather than averaging the two per-codec correlations. A metric that achieves consistently high correlation for each codec individually and for the pooled scores can be considered codec-independent within the scope of this evaluation.

\subsection{Correlation Assessment Results}\label{ssec:results}

Tables~\ref{tab:srocc-jp2k}--\ref{tab:plcc-avg} present the correlation between event stream distortion metrics and task degradation, evaluated separately per codec and over the pooled lossy operating points of both codecs, using SROCC and PLCC. The tables are divided into two row blocks; the upper block reports results for the five proposed metrics (see Section~\ref{ssec:classifmetrics}), while the lower block reports results for the six identified benchmark metrics reviewed in Section~\ref{sec:background}. All correlations are computed over the correlation pools defined in Section~\ref{ssec:correlation}: per-codec results use the pool of a single codec, while the pooled results use the merged pool of both codecs.

For the aggregation-based codec (Tables~\ref{tab:srocc-jp2k} and~\ref{tab:plcc-jp2k}), all five proposed metrics achieve SROCC~$\geq 0.80$ on all three tasks. The same behavior is observed for the PC-based codec (Tables~\ref{tab:srocc-gpcc} and~\ref{tab:plcc-gpcc}), with the lowest correlations obtained for Markedness and MCC on video reconstruction (SROCC values of~$0.817$ and~$0.833$, respectively). When the operating points of both codecs are pooled together (Tables~\ref{tab:srocc-avg} and~\ref{tab:plcc-avg}), all five proposed metrics maintain SROCC~$\geq 0.80$ across all three tasks. In contrast, the existing event-level metrics ($VQ_s$, $VQ_t$~\cite{adhuran2025lossy}, PSNR~E2E~\cite{seleem2026deep}, and ASTSM~\cite{Li:2023,stumpp2024flow}) show substantially weaker and less consistent correlation with task degradation. Under the aggregation-based codec, none of these four metrics reaches the $0.80$ SROCC threshold on any task, with values below $0.50$ on object detection and video reconstruction. Under the PC-based codec, only isolated combinations rise above the threshold (e.g., $VQ_s$ reaches SROCC of~$0.860$ on optical flow estimation), so none of the four maintains consistently high correlation across codecs and tasks. 

Frame~PSNR and Frame~MSE behave differently from the other four benchmark metrics. Considering SROCC, under the PC-based codec, both exceed the $0.80$ threshold on object detection and optical flow estimation but not on video reconstruction; under the aggregation-based codec, Frame~PSNR exceeds it only on video reconstruction, while Frame~MSE does so on video reconstruction and optical flow estimation. When the two codecs are pooled, the correlation of both metrics decreases substantially for object detection and optical flow estimation (Frame~PSNR: $0.516$ and $0.722$; Frame~MSE: $0.590$ and $0.775$), while for video reconstruction both stay above the $0.80$ threshold. The correlation of these two metrics is therefore dependent on both the task and the codec representation.

Considering all the pooled lossy operating points, the mean SROCC of the five proposed metrics across the three tasks ranges from $0.878$ for Markedness to $0.890$ for IoU, while their mean PLCC ranges from $0.858$ for Markedness to $0.872$ for Cohen's Kappa. Thus, the spread among the metrics is less than $0.02$ for both correlation measures. Moreover, all five metrics achieve at least $0.834$ SROCC and $0.801$ PLCC on every individual task (Tables~\ref{tab:srocc-avg} and~\ref{tab:plcc-avg}). The best-performing metric varies across tasks and correlation measures; consequently, no single proposed metric consistently dominates, and all five can be regarded as broadly comparable in their ability to assess compression-induced task degradation.

The central empirical evidence supporting the codec-robustness claim is that all five proposed metrics exhibit strong correlations with task performance (SROCC~$\geq 0.80$) across all three tasks, both for each codec architecture individually and when the operating points from the two codecs are pooled. In contrast, none of the six benchmark metrics meets this criterion consistently across all three tasks for either codec. The four event-level benchmark metrics remain below the threshold throughout, whereas Frame~PSNR and Frame~MSE exceed it only for specific task--codec combinations. These results suggest that the predictive power of the proposed metrics is associated primarily with how they characterize event-stream distortion rather than with the particular codec architecture.

\begin{figure*}[!t]
    \centering
    \includegraphics[width=\textwidth]{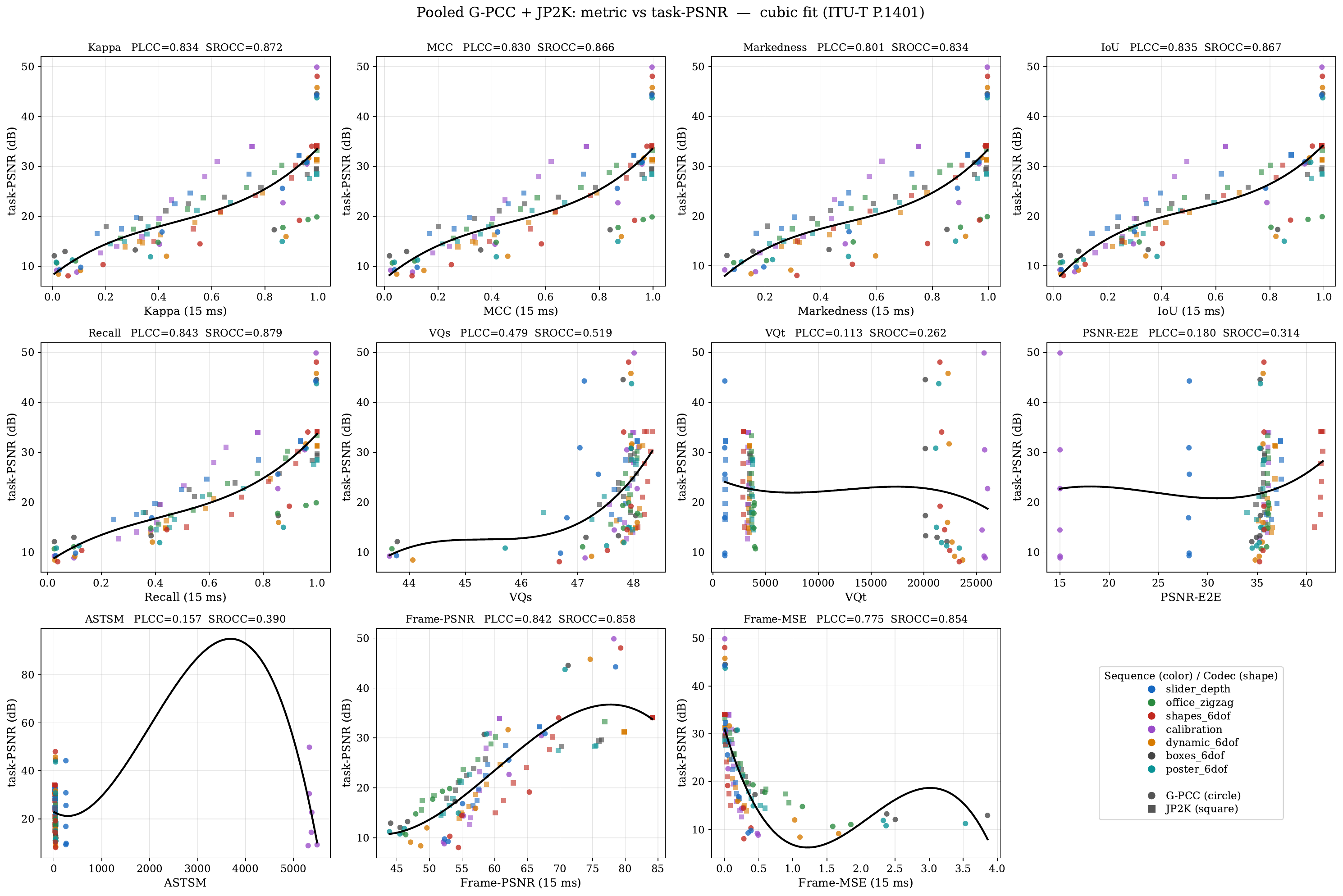}
    \caption{Video reconstruction (HyperE2VID, ECD): task score (PSNR against the raw events reference) versus each event stream distortion metric, over the pooled lossy operating points of both codecs (G-PCC- and JPEG~2000-based). Points are colored by sequence and shaped by codec; the black curve is the cubic fit described in Section~\ref{ssec:correlation}. The first five plots correspond to the proposed metrics and the remaining six to the benchmark metrics, following the same grouping as Tables~\ref{tab:srocc-jp2k}--\ref{tab:plcc-avg}.}
    \label{fig:scatter-video-all}
\end{figure*}

Figure~\ref{fig:scatter-video-all} provides a visual illustration of the correlation between metrics for the video reconstruction task. For each lossy operating point of both codecs, the task score is plotted against each of the eleven event stream metrics, together with the cubic fit described in Section~\ref{ssec:correlation} computed over all the pooled lossy points. The five proposed metrics exhibit a consistent monotonic relationship with task performance, in agreement with the correlation values reported in Tables~\ref{tab:srocc-avg} and~\ref{tab:plcc-avg}, whereas $VQ_s$, $VQ_t$, PSNR~E2E and ASTSM show visibly weaker or less consistent monotonic trends. Frame~PSNR and Frame~MSE are exceptions among the benchmark metrics, remaining competitive for video reconstruction, with pooled SROCC values of $0.858$ and $0.854$, respectively (Table~\ref{tab:srocc-avg}). Their limitations become more apparent for the other tasks: for object detection, their pooled SROCC values decrease to $0.516$ and $0.590$, respectively, while for optical flow estimation, they reach only $0.722$ and $0.775$, respectively. The corresponding scatter plots for object detection and optical flow estimation are omitted for conciseness, with the numerical results reported in Table~\ref{tab:corr-all}.

\begin{table*}[!htb]
\centering
\caption{SROCC and PLCC correlation between event stream distortion metrics and task degradation (15~ms window), evaluated separately per codec and over the pooled lossy operating points of both codecs. Bold indicates reliable prediction, defined as correlation $\geq$ 0.800.}
\label{tab:corr-all}

\begin{subtable}[t]{0.32\textwidth}
\centering
\caption{SROCC — Aggregation-based coding}
\label{tab:srocc-jp2k}
\footnotesize
\setlength{\tabcolsep}{2.5pt}
\begin{tabular*}{0.95\linewidth}{@{\extracolsep{\fill}}lccc@{}}
\toprule
\textbf{Metric} & \shortstack{\textbf{Obj.}\\\textbf{Det.}} & \shortstack{\textbf{Opt. Flow}\\\textbf{Est.}} & \shortstack{\textbf{Video}\\\textbf{Rec.}} \\
\midrule
\multicolumn{4}{l}{\textit{Proposed}} \\
Cohen's Kappa & \textbf{0.838} & \textbf{0.984} & \textbf{0.904} \\
MCC           & \textbf{0.814} & \textbf{0.984} & \textbf{0.903} \\
Markedness    & \textbf{0.935} & \textbf{0.823} & \textbf{0.903} \\
IoU           & \textbf{0.868} & \textbf{0.984} & \textbf{0.904} \\
Recall        & \textbf{0.857} & \textbf{0.972} & \textbf{0.889} \\
\midrule
\multicolumn{4}{l}{\textit{Benchmark}} \\
VQ\_s      & 0.110 & 0.676 & 0.447 \\
VQ\_t      & 0.058 & 0.285 & 0.054 \\
PSNR E2E   & 0.109 & 0.761 & 0.375 \\
ASTSM      & 0.215 & 0.585 & 0.434 \\
Frame PSNR & 0.426 & 0.775 & \textbf{0.803} \\
Frame MSE  & 0.438 & \textbf{0.802} & \textbf{0.818} \\
\bottomrule
\end{tabular*}
\end{subtable}
\hfill
\begin{subtable}[t]{0.32\textwidth}
\centering
\caption{SROCC — Point-cloud-based coding}
\label{tab:srocc-gpcc}
\footnotesize
\setlength{\tabcolsep}{2.5pt}
\begin{tabular*}{0.95\linewidth}{@{\extracolsep{\fill}}lccc@{}}
\toprule
\textbf{Metric} & \shortstack{\textbf{Obj.}\\\textbf{Det.}} & \shortstack{\textbf{Opt. Flow}\\\textbf{Est.}} & \shortstack{\textbf{Video}\\\textbf{Rec.}} \\
\midrule
\multicolumn{4}{l}{\textit{Proposed}} \\
Cohen's Kappa & \textbf{0.965} & \textbf{0.923} & \textbf{0.853} \\
MCC           & \textbf{0.940} & \textbf{0.923} & \textbf{0.833} \\
Markedness    & \textbf{0.940} & \textbf{0.965} & \textbf{0.817} \\
IoU           & \textbf{0.929} & \textbf{0.923} & \textbf{0.931} \\
Recall        & \textbf{0.983} & \textbf{0.923} & \textbf{0.894} \\
\midrule
\multicolumn{4}{l}{\textit{Benchmark}} \\
VQ\_s      & 0.714 & \textbf{0.860} & 0.647 \\
VQ\_t      & 0.235 & 0.664 & 0.284 \\
PSNR E2E   & 0.568 & 0.483 & 0.226 \\
ASTSM      & 0.702 & 0.357 & 0.067 \\
Frame PSNR & \textbf{0.906} & \textbf{0.895} & 0.792 \\
Frame MSE  & \textbf{0.896} & \textbf{0.860} & 0.788 \\
\bottomrule
\end{tabular*}
\end{subtable}
\hfill
\begin{subtable}[t]{0.32\textwidth}
\centering
\caption{SROCC — Pooled}
\label{tab:srocc-avg}
\footnotesize
\setlength{\tabcolsep}{2.5pt}
\begin{tabular*}{0.95\linewidth}{@{\extracolsep{\fill}}lccc@{}}
\toprule
\textbf{Metric} & \shortstack{\textbf{Obj.}\\\textbf{Det.}} & \shortstack{\textbf{Opt. Flow}\\\textbf{Est.}} & \shortstack{\textbf{Video}\\\textbf{Rec.}} \\
\midrule
\multicolumn{4}{l}{\textit{Proposed}} \\
Cohen's Kappa & \textbf{0.877} & \textbf{0.920} & \textbf{0.872} \\
MCC           & \textbf{0.866} & \textbf{0.920} & \textbf{0.866} \\
Markedness    & \textbf{0.862} & \textbf{0.937} & \textbf{0.834} \\
IoU           & \textbf{0.884} & \textbf{0.920} & \textbf{0.867} \\
Recall        & \textbf{0.898} & \textbf{0.876} & \textbf{0.879} \\
\midrule
\multicolumn{4}{l}{\textit{Benchmark}} \\
VQ\_s      & 0.224 & 0.499 & 0.519 \\
VQ\_t      & 0.154 & 0.036 & 0.262 \\
PSNR E2E   & 0.342 & 0.333 & 0.314 \\
ASTSM      & 0.195 & 0.274 & 0.390 \\
Frame PSNR & 0.516 & 0.722 & \textbf{0.858} \\
Frame MSE  & 0.590 & 0.775 & \textbf{0.854} \\
\bottomrule
\end{tabular*}
\end{subtable}

\vspace{6pt}

\begin{subtable}[t]{0.32\textwidth}
\centering
\caption{PLCC — Aggregation-based coding}
\label{tab:plcc-jp2k}
\footnotesize
\setlength{\tabcolsep}{2.5pt}
\begin{tabular*}{0.95\linewidth}{@{\extracolsep{\fill}}lccc@{}}
\toprule
\textbf{Metric} & \shortstack{\textbf{Obj.}\\\textbf{Det.}} & \shortstack{\textbf{Opt. Flow}\\\textbf{Est.}} & \shortstack{\textbf{Video}\\\textbf{Rec.}} \\
\midrule
\multicolumn{4}{l}{\textit{Proposed}} \\
Cohen's Kappa & \textbf{0.859} & \textbf{0.895} & \textbf{0.912} \\
MCC           & \textbf{0.838} & \textbf{0.894} & \textbf{0.910} \\
Markedness    & \textbf{0.944} & \textbf{0.885} & \textbf{0.914} \\
IoU           & \textbf{0.896} & \textbf{0.892} & \textbf{0.909} \\
Recall        & \textbf{0.940} & \textbf{0.895} & \textbf{0.891} \\
\midrule
\multicolumn{4}{l}{\textit{Benchmark}} \\
VQ\_s      & 0.113 & \textbf{0.839} & 0.462 \\
VQ\_t      & 0.034 & 0.469 & 0.172 \\
PSNR E2E   & 0.257 & 0.767 & 0.362 \\
ASTSM      & 0.369 & 0.705 & 0.415 \\
Frame PSNR & 0.727 & \textbf{0.892} & 0.787 \\
Frame MSE  & 0.787 & \textbf{0.807} & \textbf{0.823} \\
\bottomrule
\end{tabular*}
\end{subtable}
\hfill
\begin{subtable}[t]{0.32\textwidth}
\centering
\caption{PLCC — Point-cloud-based coding}
\label{tab:plcc-gpcc}
\footnotesize
\setlength{\tabcolsep}{2.5pt}
\begin{tabular*}{0.95\linewidth}{@{\extracolsep{\fill}}lccc@{}}
\toprule
\textbf{Metric} & \shortstack{\textbf{Obj.}\\\textbf{Det.}} & \shortstack{\textbf{Opt. Flow}\\\textbf{Est.}} & \shortstack{\textbf{Video}\\\textbf{Rec.}} \\
\midrule
\multicolumn{4}{l}{\textit{Proposed}} \\
Cohen's Kappa & \textbf{0.962} & \textbf{0.918} & \textbf{0.900} \\
MCC           & \textbf{0.961} & \textbf{0.920} & \textbf{0.898} \\
Markedness    & \textbf{0.954} & \textbf{0.920} & \textbf{0.874} \\
IoU           & \textbf{0.959} & \textbf{0.888} & \textbf{0.915} \\
Recall        & \textbf{0.963} & \textbf{0.912} & \textbf{0.913} \\
\midrule
\multicolumn{4}{l}{\textit{Benchmark}} \\
VQ\_s      & 0.612 & \textbf{0.976} & 0.488 \\
VQ\_t      & 0.303 & \textbf{0.834} & 0.161 \\
PSNR E2E   & 0.551 & 0.756 & 0.175 \\
ASTSM      & 0.661 & 0.763 & 0.196 \\
Frame PSNR & \textbf{0.904} & 0.680 & \textbf{0.939} \\
Frame MSE  & \textbf{0.949} & 0.745 & \textbf{0.815} \\
\bottomrule
\end{tabular*}
\end{subtable}
\hfill
\begin{subtable}[t]{0.32\textwidth}
\centering
\caption{PLCC — Pooled}
\label{tab:plcc-avg}
\footnotesize
\setlength{\tabcolsep}{2.5pt}
\begin{tabular*}{0.95\linewidth}{@{\extracolsep{\fill}}lccc@{}}
\toprule
\textbf{Metric} & \shortstack{\textbf{Obj.}\\\textbf{Det.}} & \shortstack{\textbf{Opt. Flow}\\\textbf{Est.}} & \shortstack{\textbf{Video}\\\textbf{Rec.}} \\
\midrule
\multicolumn{4}{l}{\textit{Proposed}} \\
Cohen's Kappa & \textbf{0.949} & \textbf{0.834} & \textbf{0.834} \\
MCC           & \textbf{0.946} & \textbf{0.837} & \textbf{0.830} \\
Markedness    & \textbf{0.915} & \textbf{0.857} & \textbf{0.801} \\
IoU           & \textbf{0.954} & \textbf{0.823} & \textbf{0.835} \\
Recall        & \textbf{0.947} & \textbf{0.817} & \textbf{0.843} \\
\midrule
\multicolumn{4}{l}{\textit{Benchmark}} \\
VQ\_s      & 0.521 & \textbf{0.883} & 0.479 \\
VQ\_t      & 0.262 & 0.270 & 0.113 \\
PSNR E2E   & 0.498 & 0.716 & 0.180 \\
ASTSM      & 0.479 & 0.628 & 0.157 \\
Frame PSNR & 0.771 & 0.648 & \textbf{0.843} \\
Frame MSE  & \textbf{0.837} & 0.685 & 0.775 \\
\bottomrule
\end{tabular*}
\end{subtable}
\end{table*}
\FloatBarrier
\subsection{Metric Validation for Asynchronous Event Processing} \label{ssec:async}

Many event-based vision applications apply some form of preprocessing to the event stream before the task is performed, typically converting the event stream into an intermediate representation, such as frames or voxel grids, often using pre-trained models. The three tasks evaluated so far follow this paradigm. However, some applications adopt a fundamentally different processing model based on asynchronous event-by-event computation. To assess whether the proposed metrics remain predictive under this setting, a fourth downstream task, asynchronous feature tracking, is considered. This task serves as a stress test of the proposed metrics for delay-sensitive applications. Unlike the previous three tasks, it operates directly on the decoded event stream, processing events individually as they arrive without constructing intermediate representations or relying on a trained model. Because its performance depends critically on the precise timing of individual events, it provides a more demanding evaluation of the proposed metrics for applications that exploit the native temporal resolution of event cameras. 

\subsubsection{Low-delay asynchronous feature tracking} \label{sssec:haste-task}

Asynchronous feature tracking is implemented using HASTE~\cite{Alzugaray:2020}, an event-by-event tracking algorithm that maintains a discrete set of tracking hypotheses and updates their scores incrementally with each incoming event, following the asynchronous multi-hypothesis framework of~\cite{Alzugaray:2019}. Evaluation is performed on the Event Camera Dataset (ECD)~\cite{Mueggler:2017}, the same dataset used for the video reconstruction task. The two compression pipelines considered are configured exactly as described in Sections~\ref{ssec:codec_points} and~\ref{ssec:gpcc_points}, following the same configuration used for the other three downstream tasks. The RR protocol introduced in Section~\ref{sec:task-eval} is applied unchanged, i.e., reference trajectories are first generated from the uncompressed event stream, after which tracking is repeated on every decoded stream to measure the corresponding degradation.

Tracking performance is quantified using the \emph{feature age} metric. For each seed feature, the track age is the elapsed time during which a tracked feature remains within a 2~pixel tolerance of the corresponding reference trajectory. The score reported for a sequence is the sum of the track ages over all seeds divided by the sum of the corresponding reference trajectory durations. Features whose test trajectory diverges immediately are retained in this computation and thus lost tracks lower the score rather than being discarded. The evaluation is therefore performed using six ECD sequences, those where the two compression pipelines provided decoded event streams of sufficient quality for feature tracking. The stream distortion metrics are computed as previously and thus independently of the coding pipeline.

Since HASTE performs only tracking and does not include a feature detector, the initial feature locations must be supplied externally, following the evaluation protocol of \cite{Alzugaray:2020}. For each sequence, up to 30 Harris corners are detected on the first APS intensity frame provided by the DAVIS240C sensor. These APS images are not part of either compression pipeline and are used only to initialise tracking. The same set of feature locations is then reused for every operating point and for both codecs, ensuring that differences in feature age reflect only the quality of the decoded event stream rather than variations in feature initialisation. The tracker is the HasteCorrelation$^\star$ variant of~\mbox{\cite{Alzugaray:2020}}, which updates each hypothesis score incrementally, built on the multi-hypothesis framework of~\mbox{\cite{Alzugaray:2019}}.

\begin{figure}[!t]
    \centering
    \begin{subfigure}{0.65\linewidth}
        \centering
        \includegraphics[width=\linewidth]{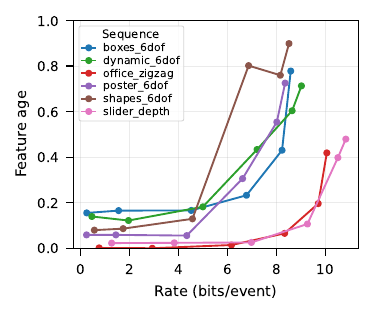}
        \subcaption{Point-cloud-based coding.}
    \end{subfigure}
    
    \vspace{4pt}
    
    \begin{subfigure}{0.65\linewidth}
        \centering
        \includegraphics[width=\linewidth]{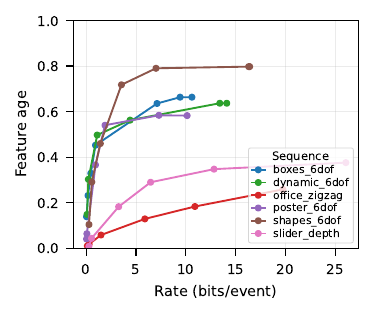}
        \subcaption{Aggregation-based codec.}
    \end{subfigure}
    \caption{Rate-utility curves for asynchronous feature tracking (HASTE, ECD), evaluated over the operating points defined in Sections~\ref{ssec:codec_points} and~\ref{ssec:gpcc_points} under the RR protocol. Each curve corresponds to one sequence. As in Fig.~\ref{fig:rd-jp2k}, each curve of the aggregation-based codec shows the convex hull of the operating points.}
    \label{fig:rd-haste}
\end{figure}

Fig.~\ref{fig:rd-haste} shows the resulting rate--utility curves for the asynchronous feature tracking task under both codecs. As for the other three tasks, tracking performance (task score) degrades as the bitrate decreases, at a pace that depends on the sequence and on the codec, although the degradation patterns differ. For the PC-based codec, the feature age is largely preserved down to a rate that varies with the sequence before exhibiting a sharp drop, whereas for the aggregation-based codec, the feature age gradually improves with increasing bitrate before reaching saturation at the higher rates. In contrast to the previous three tasks, however, neither codec recovers the reference tracking performance, even at the highest bitrate operating point. This residual degradation is consistent with the behavior anticipated in Section~\ref{ssec:codec-assessment}: even at their highest-quality settings, both codecs introduce irreversible event-level distortions through temporal aggregation or mapping to quantized point-cloud coordinates. These distortions are clear here because the tracker processes events individually, whereas the other three tasks re-bin the decoded streams before inference and are therefore less sensitive to them.

Tables~\ref{tab:srocc-haste} and~\ref{tab:plcc-haste} report the correlation scores (as described in Section~\ref{ssec:correlation}) between the eleven event stream distortion metrics and the feature age metric. Although the correlations are lower than those obtained for the previous three tasks, the proposed metrics remain the strongest predictors in terms of rank correlation. On the pooled operating points, the five proposed metrics reach SROCC values between $0.788$ and $0.808$, outperforming all benchmark metrics, with Frame~PSNR ($0.787$) being the closest competitor. Four of the five proposed metrics exceed the $0.80$ threshold defined in Section~\ref{ssec:correlation} and Markedness falls just below it. For the point cloud codec alone, however, none of the five metrics reaches the threshold. The ordering changes under PLCC, where pooled Frame~PSNR ($0.818$) surpasses all five proposed metrics ($0.777$--$0.795$). In conclusion, the proposed metrics remain the most reliable predictors of task degradation under rank correlation, even for tasks that processes the decoded event stream directly on an event-by-event basis.

\begin{table}[!htb]
\centering
\caption{SROCC and PLCC correlation for asynchronous feature tracking (15~ms window), reported separately per codec and over the pooled lossy operating points of both codecs. Bold indicates reliable prediction, defined as correlation $\geq$ 0.800.}
\label{tab:haste-all}

\begin{subtable}[t]{\linewidth}
\centering
\caption{SROCC}
\label{tab:srocc-haste}
\footnotesize
\setlength{\tabcolsep}{6pt}
\begin{tabular}{lccc}
\toprule
\textbf{Metric} & \textbf{PC-based} & \textbf{Aggr.-based} & \textbf{Pooled} \\
\midrule
\multicolumn{4}{l}{\textit{Proposed}} \\
Cohen's Kappa & 0.761 & \textbf{0.813} & \textbf{0.807} \\
MCC           & 0.765 & \textbf{0.813} & \textbf{0.807} \\
Markedness    & 0.775 & 0.798 & 0.788 \\
IoU           & 0.792 & 0.787 & \textbf{0.801} \\
Recall        & 0.793 & \textbf{0.800} & \textbf{0.808} \\
\midrule
\multicolumn{4}{l}{\textit{Benchmark}} \\
VQ\_s      & 0.604 & 0.523 & 0.535 \\
VQ\_t      & 0.570 & 0.504 & 0.088 \\
PSNR E2E   & 0.444 & 0.185 & 0.269 \\
ASTSM      & 0.242 & 0.375 & 0.279 \\
Frame PSNR & 0.740 & 0.789 & 0.787 \\
Frame MSE  & 0.701 & 0.789 & 0.783 \\
\bottomrule
\end{tabular}
\end{subtable}

\vspace{6pt}

\begin{subtable}[t]{\linewidth}
\centering
\caption{PLCC}
\label{tab:plcc-haste}
\footnotesize
\setlength{\tabcolsep}{6pt}
\begin{tabular}{lccc}
\toprule
\textbf{Metric} & \textbf{PC-based} & \textbf{Aggr.-based} & \textbf{Pooled} \\
\midrule
\multicolumn{4}{l}{\textit{Proposed}} \\
Cohen's Kappa & \textbf{0.869} & \textbf{0.804} & 0.795 \\
MCC           & \textbf{0.869} & \textbf{0.804} & 0.794 \\
Markedness    & \textbf{0.876} & \textbf{0.802} & 0.777 \\
IoU           & \textbf{0.868} & 0.791 & 0.785 \\
Recall        & \textbf{0.856} & 0.782 & 0.795 \\
\midrule
\multicolumn{4}{l}{\textit{Benchmark}} \\
VQ\_s      & 0.556 & 0.602 & 0.555 \\
VQ\_t      & 0.508 & 0.550 & 0.231 \\
PSNR E2E   & 0.370 & 0.525 & 0.404 \\
ASTSM      & 0.298 & 0.509 & 0.417 \\
Frame PSNR & \textbf{0.864} & \textbf{0.804} & \textbf{0.818} \\
Frame MSE  & 0.777 & 0.774 & 0.750 \\
\bottomrule
\end{tabular}
\end{subtable}
\end{table}
\subsubsection{Influence of the metric temporal window} \label{sssec:tw}

All results reported so far were obtained with the temporal window used to compute the metrics, $t_w$, fixed at 15~ms, as introduced in Section~\ref{sec:metrics}. This window is independent of both the aggregation interval used by the codec and any temporal representation formed internally by the downstream task. Therefore, it is important to understand how it affects the ability of the metrics to predict task degradation, and whether the reported correlations depend on the specific value chosen. This is examined here by varying $t_w$ over a wide range and observing its effect on the correlation with task performance.

Video reconstruction and asynchronous feature tracking provide an ideal setting for this analysis because they are evaluated on the same six ECD sequences as in the previous Section, using identical operating points and decoded event streams, differing only in the downstream task. The two tasks, however, operate at markedly different temporal scales: video reconstruction integrates events over tens of milliseconds, whereas feature tracking processes each event individually as it arrives. Consequently, the observed differences in the optimal value of $t_w$ can be attributed mainly to the temporal characteristics of the downstream task rather than to the compressed data. Cohen's Kappa is therefore recomputed for ten values of \mbox{$t_w$} ranging from 1 to 75~ms for both codecs.

\begin{figure}[!t]
    \centering
    \begin{subfigure}{0.65\linewidth}
        \centering
        \includegraphics[width=\linewidth]{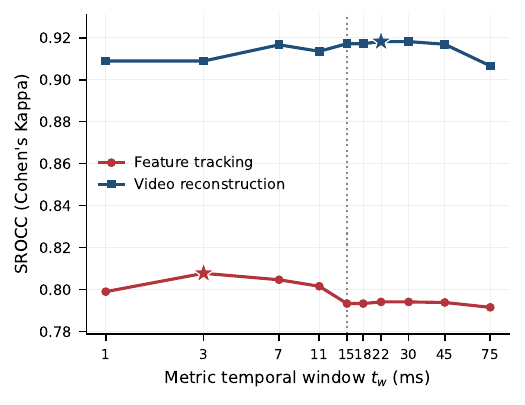}
        \subcaption{Point cloud-based codec.}
    \end{subfigure}
    
    \vspace{4pt}
    
    \begin{subfigure}{0.65\linewidth}
        \centering
        \includegraphics[width=\linewidth]{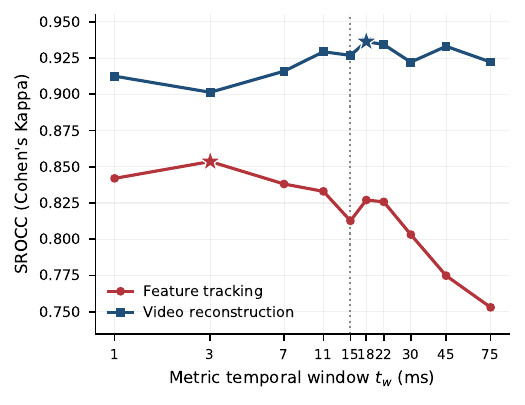}
        \subcaption{Aggregation-based codec.}
    \end{subfigure}
    \caption{SROCC between Cohen's Kappa and the task score as a function of the metric temporal window $t_w$, for the point cloud-based codec (top) and the aggregation-based codec (bottom); red marks feature tracking and blue video reconstruction. Both tasks use the same six ECD sequences, operating points and decoded event streams, so the downstream task is the only variable within each plot. Stars mark each curve's maximum; the dotted line marks the 15~ms setting defined in Section~\ref{sec:metrics}.}
    \label{fig:tw}
\end{figure}

Fig.~\ref{fig:tw} reports SROCC as a function of $t_w$ for Cohen's Kappa, for both tasks and both codecs, without the cubic mapping of Section~\ref{ssec:correlation} so that the correlations reflect the metric as it is computed. For asynchronous feature tracking, the highest correlations are consistently obtained with short temporal windows, and SROCC decreases as $t_w$ increases. Under the point cloud-based codec, it decreases from $0.81$ at 3~ms to $0.79$ at 75~ms, while under the aggregation-based codec the reduction is more pronounced, from $0.85$ at 3~ms to $0.75$ at 75~ms. Video reconstruction exhibits the opposite behaviour, being unaffected or marginally favoured by intermediate windows. The highest SROCC values are $0.918$ at 22 ms for the point cloud-based codec and $0.936$ at 18 ms for the aggregation-based codec, compared with $0.907$ and $0.922$, respectively, at 75 ms. These results indicate that the preferred temporal window is determined primarily by the downstream task rather than by the codec. Short windows are favoured by asynchronous feature tracking, which depends on the precise timing of individual events, whereas video reconstruction benefits from intermediate windows that better match its temporal integration process. The same ordering is observed for both codecs, although the dependence on $t_w$ is stronger for the aggregation-based codec, where the maxima for the two tasks occur at clearly separated windows.

The temporal window is therefore not an intrinsic property of the proposed metrics but a task-dependent parameter that can be adjusted to match the temporal characteristics of the target application. In particular, more delay-sensitive tasks can be accommodated simply by reducing $t_w$ without modifying the metric formulation itself. The 15~ms window adopted in Section~\ref{sec:metrics} lies close to the optimum for video reconstruction and remains within $0.05$ SROCC of the maximum correlation for every task-codec combination. The largest gap occurs for feature tracking under the aggregation-based codec, where a 15~ms window yields a SROCC of $0.81$, compared with the maximum of $0.85$ achieved at 3~ms. 

\section{Conclusions}\label{sec:conclusions}

This paper presents two event stream compression pipelines, an aggregation-based solution using JPEG~2000 and a frame-free point cloud-based solution using G-PCC. It then proposes a comprehensive task-driven evaluation framework that links event stream distortion to application-level performance under a reference-relative protocol. This framework enables systematic assessment of the impact of lossy compression across several downstream computer vision tasks, including video reconstruction, object detection, optical flow estimation and asynchronous feature tracking.

More importantly, this work is the first to apply classification-based and anomaly-detection distortion metrics to event compression, exploiting their suitability to the extreme class imbalance of the errors on the decoded streams. The proposed metrics, Cohen's Kappa, MCC, Markedness, IoU, and Recall, are systematically compared against six existing event stream benchmarks. The correlation analysis shows that all five proposed metrics predict task performance reliably, reaching SROCC and PLCC values above $0.80$ on every task-codec combination and on the pooled scores across both codecs, while none of the six existing benchmarks achieve a similar level of performance in general. This behaviour is further validated on the more demanding asynchronous feature tracking task, where the proposed metrics again achieve the highest rank correlation on the pooled operating points (SROCC of $0.788$--$0.808$), despite the task's much stricter temporal requirements. The central contribution is therefore not merely a set of metrics but an enabling capability for codec design. Because the proposed metrics predict task degradation consistently across both an aggregation-based and a point cloud-based codec, they can serve as codec-independent performance indicators that replace computationally expensive task-specific evaluation during rate-distortion optimization, enabling more practical optimization of future event stream compression systems.

Future work includes the design of image and point cloud-based codecs that use the proposed metrics directly for rate-distortion optimization, and the refinement of the metrics themselves by adapting the temporal aggregation window dynamically to the sparsity of the content and to the characteristics of the downstream task, enabling codec fine-tuning for the target application.

\bibliographystyle{IEEEtran}
\bibliography{refs}

\vspace{-10\baselineskip}

\begin{IEEEbiography}[{\includegraphics[width=1in,height=1.5in,clip,keepaspectratio]{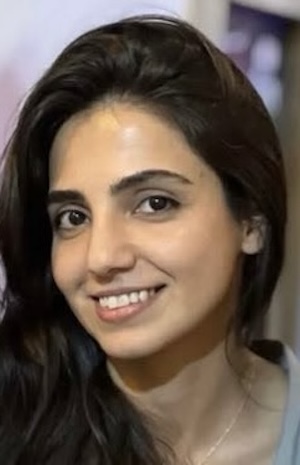}}]{Zahra Rezaee} is currently pursuing the M.Sc. degree in Computer Science and Engineering with Instituto Superior Técnico (IST), Universidade de Lisboa, Lisbon, Portugal. She is also a Research Assistant with the Instituto de Telecomunicações, working on deep learning-based compression of event-based and neuromorphic vision data. Her research interests include machine learning, computer vision, image and video compression, and event-based vision.

\end{IEEEbiography}
\vskip 0pt plus -1fil
\vspace{-3\baselineskip}

\begin{IEEEbiography}[{\includegraphics[width=1in,height=2.5in,clip,keepaspectratio]{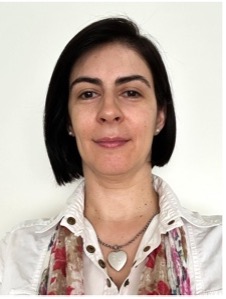}}]{Catarina Brites} (Member, IEEE) received the E.E., M.Sc., and Ph.D. degrees in electrical and computer engineering from Instituto Superior Técnico, Universidade Técnica de Lisboa, Lisbon, Portugal, in 2003, 2005, and 2011, respectively. She is currently an Assistant Professor with the Department of Information Science and Technology, ISCTE-Instituto Universitário de Lisboa, and a member of the Multimedia Signal Processing Group of Instituto de Telecomunicações, Lisbon, Portugal. She has authored more than 75 international journals and conference papers. Her current research interests include 2D/3D video processing and coding, event data coding, plenoptic imaging representations, and machine learning. She is or has been an Associate Editor of IEEE Open Journal of Signal Processing and IEEE Transactions on Image Processing and serves as a technical program committee member for several international journals and conferences in the multimedia signal processing field.

\end{IEEEbiography}
\vskip 0pt plus -1fil
\vspace{-3\baselineskip}

\begin{IEEEbiography}[{\includegraphics[width=1in,height=1.25in,clip,keepaspectratio]{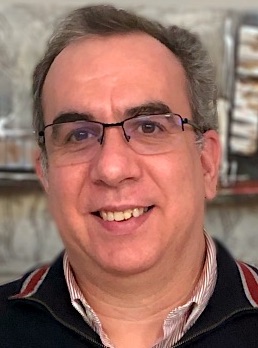}}]{Joao Ascenso} (Senior Member, IEEE)
received the E.E., M.Sc., and Ph.D. degrees in electrical and computer engineering from the Instituto Superior Técnico (IST), Universidade Técnica de Lisboa, Lisbon, Portugal, in 1999, 2003, and 2010, respectively. He is currently an Associate Professor with the Department of Electrical and Computer Engineering, IST, and a member of the Instituto de Telecomunicações. He has published more than 150 papers in international conferences and journals. His current research interests include visual coding, quality assessment, coding and processing of 3D visual representations, coding for machines, super-resolution, denoising among others. He was an associate editor of IEEE Transactions on Image Processing, IEEE Signal Processing Letters and IEEE Transactions on Multimedia and is currently senior area editor of IEEE Transactions on Circuits and Systems for Video Technology. He received three Best Paper Awards at PCS 2015, ICME 2020, and MMSP 2024 and has served as Technical Program Chair and in other organizing committees of major international conferences, including IEEE ICIP, PCS, EUVIP, ICME, MMSP and ISM. 
\end{IEEEbiography}

\EOD

\end{document}